\documentclass[final]{cvpr}

\usepackage{times}
\usepackage{epsfig}
\usepackage{graphicx}
\usepackage{amsmath}
\usepackage{amssymb}

\usepackage[normalem]{ulem}
\usepackage{multirow,booktabs}
\usepackage{footnote}
\usepackage{kotex}
\usepackage{array}
\usepackage{adjustbox}
\usepackage[table]{xcolor}
\usepackage[flushleft]{threeparttable}
\usepackage{subcaption}

\usepackage{enumitem}
\usepackage{bm}

\definecolor{Highlight}{HTML}{39b54a}  
\newcommand{\hl}[1]{\textcolor{Highlight}{#1}}

\newcommand{\PAR}[1]{\vskip4pt \noindent {\bf #1~}} 

\usepackage[pagebackref=true,breaklinks=true,colorlinks,bookmarks=false]{hyperref}

\def\cvprPaperID{****} 
\def\confYear{CVPR 2021}

\begin{document}

\title{What If We Only Use Real Datasets for Scene Text Recognition? \\
Toward Scene Text Recognition With Fewer Labels}

\author{Jeonghun Baek\qquad\qquad 
Yusuke Matsui\qquad\qquad 
Kiyoharu Aizawa\vspace{1mm}\\
The University of Tokyo \\
{\tt\small \{baek, matsui, aizawa\}@hal.t.u-tokyo.ac.jp}
}

\maketitle

\begin{abstract}
Scene text recognition (STR) task has a common practice:
All state-of-the-art STR models are trained on large synthetic data.
In contrast to this practice, training STR models only on fewer real labels (STR with fewer labels) is important when we have to train STR models without synthetic data: for handwritten or artistic texts that are difficult to generate synthetically and for languages other than English for which we do not always have synthetic data.
However, there has been implicit common knowledge that training STR models on real data is nearly impossible because real data is insufficient.
We consider that this common knowledge has obstructed the study of STR with fewer labels.
In this work, we would like to reactivate STR with fewer labels by disproving the common knowledge.
We consolidate recently accumulated public real data and show that we can train STR models satisfactorily only with real labeled data.
Subsequently, we find simple data augmentation to fully exploit real data.
Furthermore, we improve the models by collecting unlabeled data and introducing semi- and self-supervised methods.
As a result, we obtain a competitive model to state-of-the-art methods.
To the best of our knowledge, this is the first study that 1) shows sufficient performance by only using real labels and 2) introduces semi- and self-supervised methods into STR with fewer labels.
Our code and data are available: \url{https://github.com/ku21fan/STR-Fewer-Labels}.
\end{abstract}

\section{Introduction}\label{sec:introduction} 
Reading text in natural scenes is generally divided into two tasks: detecting text regions in scene images and recognizing the text in the regions.
The former is referred to as scene text detection (STD), and the latter as scene text recognition (STR).
Since STR can serve as a substitute for manual typing performed by humans, we frequently employ STR for various purposes:
translation by recognizing foreign languages, street sign recognition for autonomous driving, various card recognition to input personal information, etc.
Unlike optical character recognition (OCR), which focuses on reading texts in cleaned documents, STR also addresses irregular cases in our lives, such as curved or perspective texts, occluded texts, texts in low-resolution images, and texts written in difficult font.

To address these irregular cases, prior works have developed STR models comprising deep neural networks.
For example, to address curved or perspective texts, image transformation modules have been proposed to normalize them into horizontal images~\cite{ASTER,ESIR,ScRN}.
Qiao \etal~\cite{SE-ASTER} has integrated a pretrained language model into STR models to recognize occluded text.
Wang \etal~\cite{wang2020scene-SR-ECCV} and Mou \etal~\cite{plugnet} have introduced a super-resolution module into STR models to handle low-resolution images.

While prior works have improved STR models, the study of training STR models only on fewer real labels (STR with fewer labels) is insufficient.
After emerging large synthetic data~\cite{MJSynth} in 2014, the study of STR with fewer labels has decreased.
All state-of-the-art methods use large synthetic data to train STR models instead of sole real data~\cite{CRNN,ASTER,SAR,ESIR,Mask-textspotter,TRBA,ScRN,DAN,textscanner,SRN,scatter,SE-ASTER,robustscanner,plugnet}.
Implicit common knowledge has been made; \textit{training STR models only on real data results in low accuracy because the amount of real data is very small.}
This common knowledge may have hindered studies on STR with fewer labels.

STR with fewer labels is important when we have to train STR models without synthetic data.
In practical applications, generating synthetic data close to real data can be difficult depending on the target domain, such as handwritten text or artistic text.
In the other case, when we have to recognize languages other than English, there are not always synthetic data for them.
Generating appropriate synthetic data for them is difficult for those who do not know target languages.

\begin{figure}[t]
\includegraphics[width=0.95\linewidth]{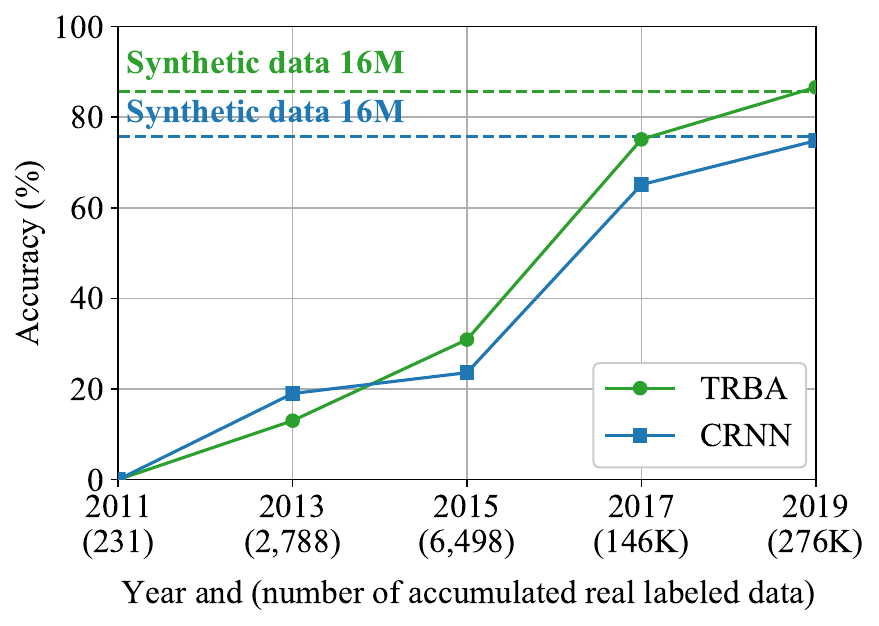}
    \centering
    \vspace{-3mm}
  \caption{Accuracy vs. number of accumulated real labeled data.
Every two years, public real data has been accumulated.
In our experiments, we find that accuracy obtained using real data approaches that obtained using synthetic data, along with increment of real data.
  CRNN~\cite{CRNN} and TRBA~\cite{TRBA} are VGG-based and ResNet-based STR models, respectively.}
  \label{fig:data_increment}
  \vspace{-2mm}
\end{figure}

In this paper, we would like to reactivate STR with fewer labels for such cases.
As a first step, we disprove the common knowledge by showing that we can train STR models satisfactorily only with real labels.
This is not previously feasible.
Because the real data was small, STR models trained on real data had low accuracy, as shown in Figure~\ref{fig:data_increment}.
However, the public real data are accumulated every two years. 
We consolidate accumulated real data~(276K), and find that the accuracy of STR models~\cite{CRNN,TRBA} trained on them is close to that of synthetic data~(16M).
Namely, we can train STR models only on real data instead of synthetic data.
It is high time to change the prevalent perspective from \textit{``We don't have enough real data to train STR models''} to \textit{``We have enough real data to train STR models''}.
It is also a good time to study STR with fewer labels.

To improve STR with fewer labels, we find simple yet effective data augmentations to fully exploit real data.
In addition, we collect unlabeled data and introduce a semi- and self-supervised framework into the STR. 
With extensive experiments, we analyze the contribution of them and demonstrate that we can obtain a competitive model to state-of-the-art methods by only using real data.
Furthermore, we investigate if our method is also useful when we have both synthetic and real data.

\section{Common Practice in STR Dataset}\label{sec:common}
According to a benchmark study~\cite{TRBA}, obtaining enough real data is difficult because of the high labeling cost.
Thus, STR models are generally trained on large synthetic data instead of real data.
Real data has been used for evaluation.

\subsection{Synthetic Datasets for Training}
There are two major synthetic datasets.

\begin{figure}[t]
\centering
    \begin{subfigure}{0.49\linewidth} \centering
     \includegraphics[width=0.97\linewidth, height=2.3cm]{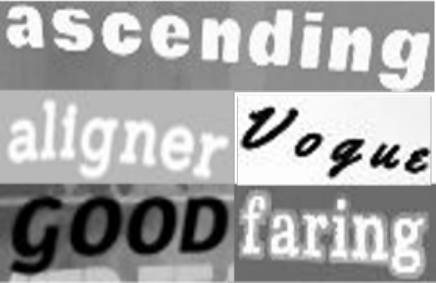}
     \caption{MJ word boxes}\label{fig:MJ}
    \end{subfigure}
    \begin{subfigure}{0.49\linewidth} \centering
     \includegraphics[width=0.97\linewidth, height=2.3cm]{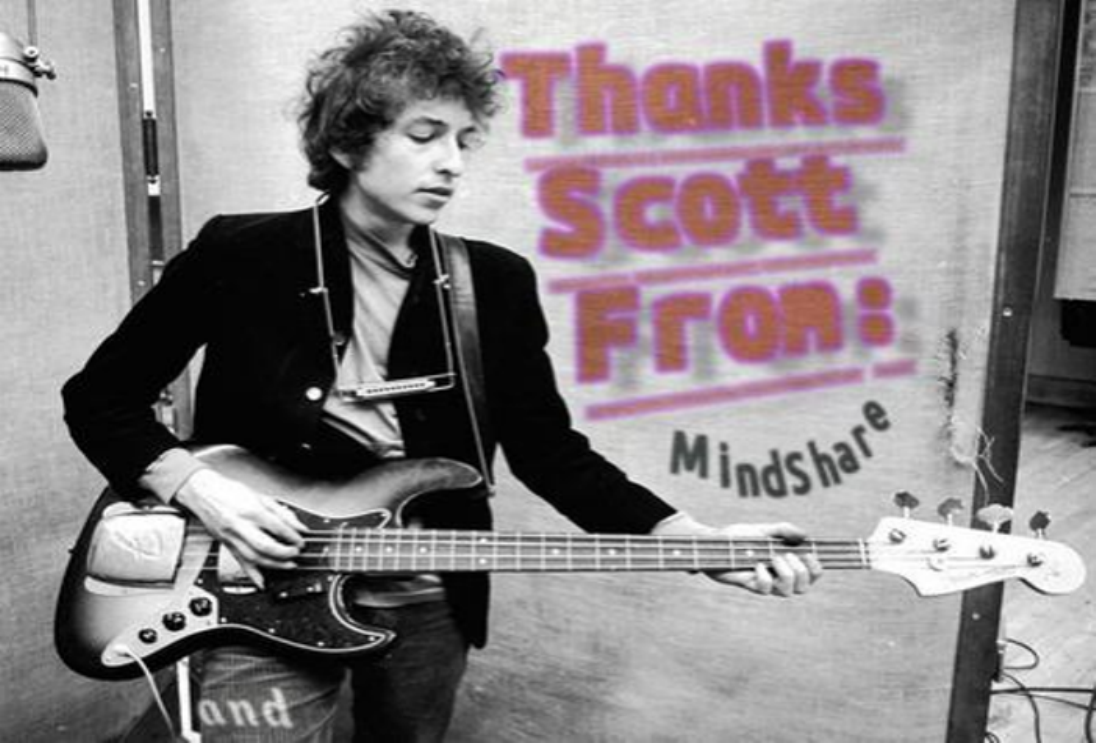}
     \caption{ST scene image}\label{fig:ST}
    \end{subfigure}
\vspace{-2mm}
\caption{Examples of two major synthetic datasets.}
\label{fig:synth}
\vspace{-2mm}
\end{figure}

\PAR{MJSynth}~\textbf{(MJ)}~\cite{MJSynth} is generated for STR, and it contains 9M word boxes.
Each word is generated from a 90K English lexicon and over 1,400 Google Fonts, as shown in Figure~\ref{fig:MJ}. 

\PAR{SynthText}~\textbf{(ST)}~\cite{SynthText} is originally generated for scene text detection.
The texts are rendered onto scene images, as shown in Figure~\ref{fig:ST}.
For STR, we crop the texts in scene images and use them for training.
ST has 7M word boxes.

\subsection{Real Benchmark Datasets for Evaluation}\label{sec:benchmark_data}
Six real datasets have been used to evaluate STR models.
\PAR{Street View Text}~\textbf{(SVT)}~\cite{SVT} is collected from Google Street View, and contains texts in street images.
It contains 257 images for training and 647 images for evaluation.

\PAR{IIIT5K-Words}~\textbf{(IIIT)}~\cite{IIIT5K} is crawled from Google image searches with query words such as ``billboards'' and ``movie posters.''
It contains 2,000 images for training and 3,000 images for evaluation.

\PAR{ICDAR2013}~\textbf{(IC13)}~\cite{IC13} is created for the ICDAR 2013 Robust Reading competition.
It contains 848 images for training and 1,015 images for evaluation.

\PAR{ICDAR2015}~\textbf{(IC15)}~\cite{IC15} is collected by people who wear Google Glass, and thus, many of them contain perspective texts and some of them are blurry.
It contains 4,468 images for training and 2,077 images for evaluation.
    
\PAR{SVT Perspective}~\textbf{(SP)}~\cite{SVTP} is collected from Google Street View, similar to SVT. 
Unlike SVT, SP contains many perspective texts.
It contains 645 images for evaluation.

\PAR{CUTE80}~\textbf{(CT)}~\cite{CUTE80} is collected for curved text. 
The images are captured by a digital camera or collected from the Internet. 
It contains 288 cropped images for evaluation.

They are generally divided into regular (SVT, IIIT, IC13) and irregular (IC15, SP, CT) datasets.
The former mainly contains horizontal texts, while the latter mainly contains perspective or curved texts.

\begin{figure*}[t]
\centering
    \begin{subfigure}{0.196\linewidth} \centering
     \includegraphics[width=0.97\linewidth, height=2.3cm]{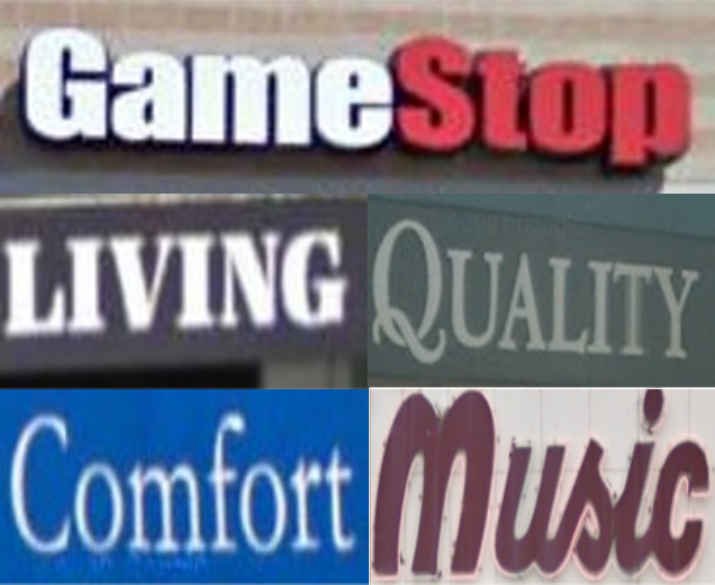}
     \vspace{-1mm}
     \caption{Year 2011}\label{fig:2011}
    \end{subfigure}
    \begin{subfigure}{0.196\linewidth} \centering
     \includegraphics[width=0.97\linewidth, height=2.3cm]{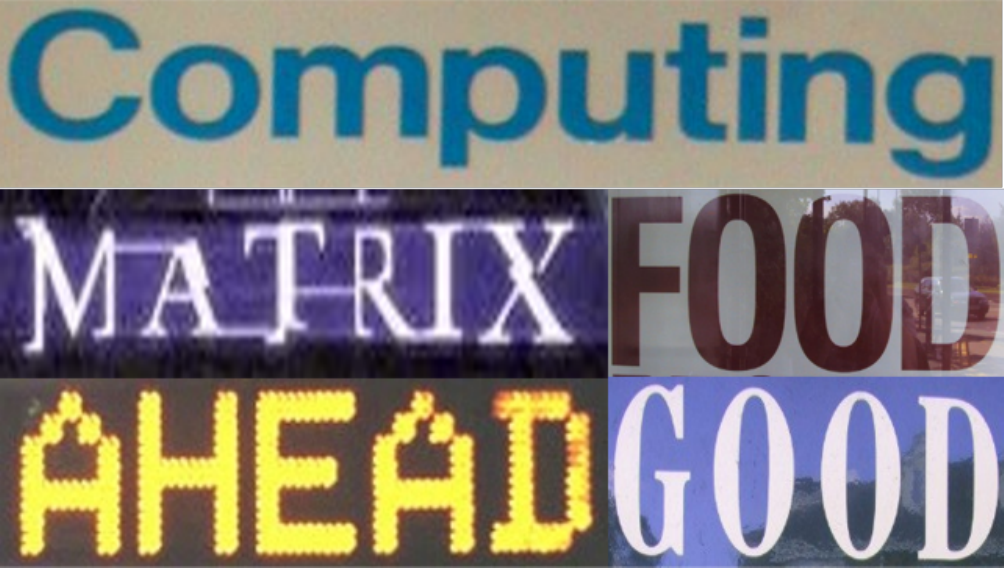}
     \vspace{-1mm}
     \caption{Year 2013}\label{fig:2013}
    \end{subfigure}
    \begin{subfigure}{0.196\linewidth} \centering
     \includegraphics[width=0.97\linewidth, height=2.3cm]{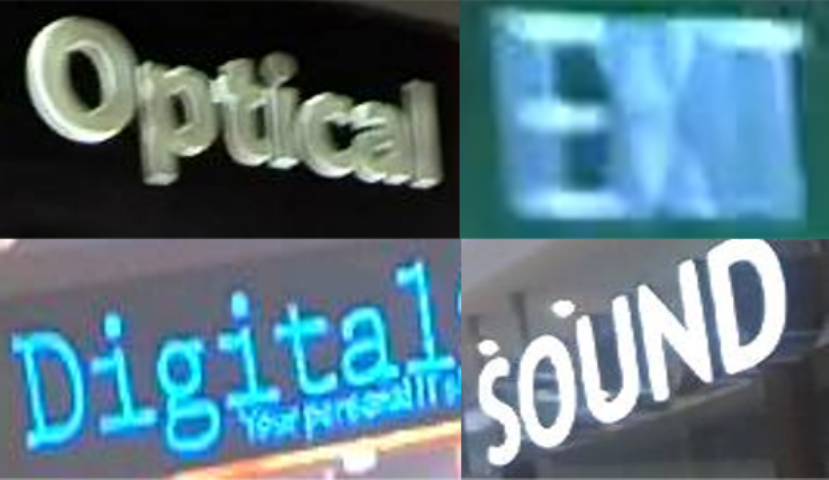}
     \vspace{-1mm}
     \caption{Year 2015}\label{fig:2015}
    \end{subfigure}
    \begin{subfigure}{0.196\linewidth} \centering
     \includegraphics[width=0.97\linewidth, height=2.3cm]{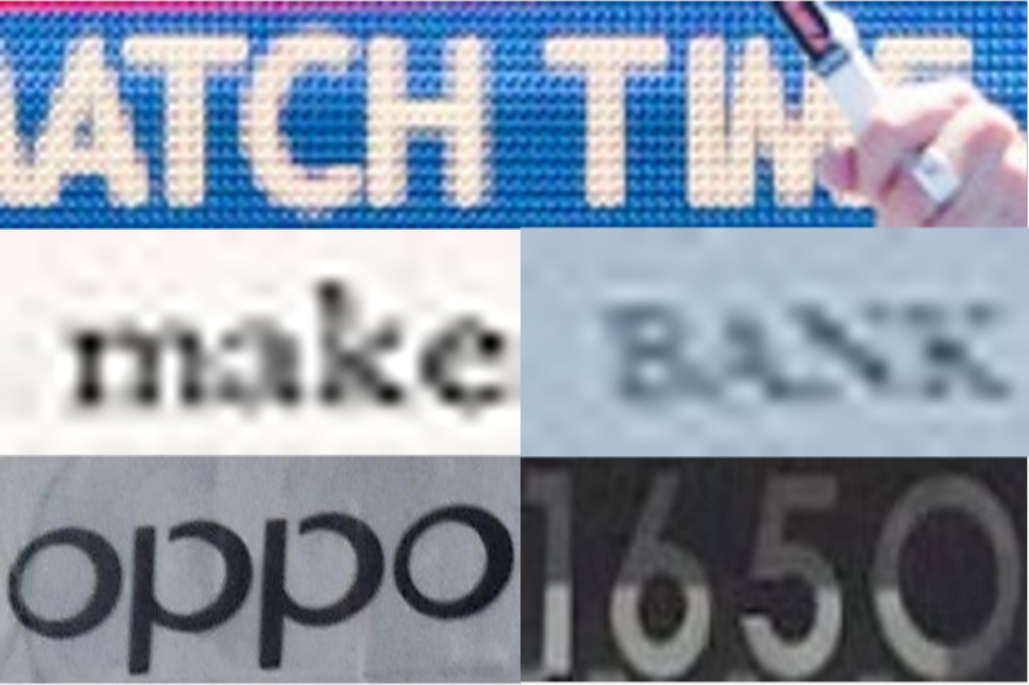}
     \vspace{-1mm}
     \caption{Year 2017}\label{fig:2017}
    \end{subfigure}
    \begin{subfigure}{0.196\linewidth} \centering
     \includegraphics[width=0.97\linewidth, height=2.3cm]{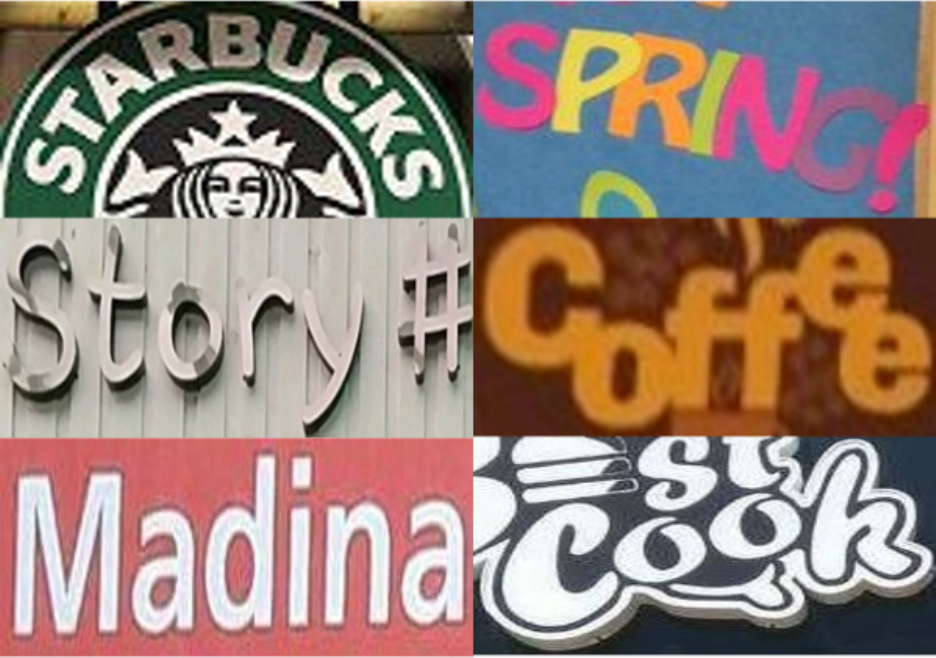}
     \vspace{-1mm}
     \caption{Year 2019}\label{fig:2019}
    \end{subfigure}
\vspace{-3mm}
\caption{Examples of accumulated real labeled data. 
More examples are provided in the supplementary materials.} 
\label{fig:dataset}
\vspace{-2mm}
\end{figure*}

\section{Consolidating Public Real Datasets}\label{sec:consoli}
Recently, public real data has been sufficiently accumulated to train STR models, as shown in Figure~\ref{fig:data_increment}. 
We consolidate the training set of public real datasets from 2011 to 2019.
Table~\ref{tab:dataset} lists datasets.
Figure~\ref{fig:dataset} shows the examples of word boxes.
Before using the original data directly for training, we conduct some preprocessing on datasets for our task.
We summarize the processes in \S\ref{sec:preprocessing}, and details are in the supplementary materials.

\subsection{Real Labeled Datasets Have Increased}\label{sec:real_label_increment}
Recently, many irregular texts are accumulated as shown in Year 2015, 2017, and 2019 in Figure~\ref{fig:dataset}. 
They can make STR models more robust.
Many real labeled datasets are released from ICDAR competitions: IC13, IC15, RCTW, ArT, LSVT, MLT19, and ReCTS (7 of 11 datasets in Table~\ref{tab:dataset}).
ICDAR competitions are held every two years, and real labeled datasets have also increased in number.
We summarize real labeled datasets for every two years.

\begin{enumerate}[label=(\alph*)]
    \item Year~2011 (SVT) and (b) Year~2013 (IIIT, IC13): Most of images are horizontal texts in the street.
    \vspace{-2mm}
\end{enumerate}

\begin{enumerate}[label=(\alph*), start=3]
    \item Year~2015 (IC15): Images captured by Google Glass under movement of the wearer, and thus many are perspective texts, blurry, or low-resolution images.
    \vspace{-2mm}
    
    \item Year~2017 (COCO, RCTW, Uber): 
    
        \textbf{COCO-Text}~\textbf{(COCO)}~\cite{COCO} is created from the MS COCO dataset~\cite{MSCOCO}. 
        As the MS COCO dataset is not intended to capture text, COCO contains many occluded or low-resolution texts.
        
        \textbf{RCTW}~\cite{RCTW} is created for \textbf{R}eading \textbf{C}hinese \textbf{T}ext in the \textbf{W}ild competition. 
        Thus many are Chinese text.
        
        \textbf{Uber-Text}~\textbf{(Uber)}~\cite{Uber} is collected from Bing Maps Streetside.
        Many are house number, and some are text on signboards.
        \vspace{-2mm}

    \item Year~2019 (ArT, LSVT, MLT19, ReCTS): 
    
        \textbf{ArT}~\cite{ArT} is created to recognize \textbf{Ar}bitrary-shaped \textbf{T}ext.
        Many are perspective or curved texts.
        It also includes Totaltext~\cite{totaltext} and CTW1500~\cite{CTW1500}, which contain many rotated or curved texts.
        
        \textbf{LSVT}~\cite{LSVT,sun2019LSVT} is a \textbf{L}arge-scale \textbf{S}treet \textbf{V}iew \textbf{T}ext dataset, collected from streets in China, and thus many are Chinese text.
        
        \textbf{MLT19}~\cite{MLT19} is created to recognize \textbf{M}ulti-\textbf{L}ingual \textbf{T}ext.
        It consists of seven languages: Arabic, Latin, Chinese, Japanese, Korean, Bangla, and Hindi.
        
        \textbf{ReCTS}~\cite{ReCTS} is created for the \textbf{Re}ading \textbf{C}hinese \textbf{T}ext on \textbf{S}ignboard competition.
        It contains many irregular texts arranged in various layouts or written with unique fonts.
\end{enumerate}

\begin{table}[t] 
  \tabcolsep=0.13cm
    \begin{center}
        \begin{adjustbox}{width=1\linewidth}
        \begin{tabular}{@{}llllrr@{}}
            \toprule
            & & & & \multicolumn{2}{c}{\# of word boxes} \\ \cmidrule(l){5-6}
            \multicolumn{2}{l}{Dataset} & Conf. & Year & Original &                   Processed \\
            \midrule
            \multicolumn{4}{l}{\textbf{Real labeled datasets (Real-L)}} \\ 
            ~~(a) & SVT~\cite{SVT}  & ICCV & 2011 & 257 & 231\\
            \arrayrulecolor{lightgray}\hline\arrayrulecolor{black}
            ~~\multirow{2}{*}{(b)} & IIIT~\cite{IIIT5K} & BMVC & 2012 & 2,000 & 1,794 \\
             & IC13~\cite{IC13} & ICDAR & 2013 & 848 & 763 \\
            \arrayrulecolor{lightgray}\hline\arrayrulecolor{black}
            ~~(c) & IC15~\cite{IC15} & ICDAR & 2015 & 4,468 & 3,710 \\ 
            \arrayrulecolor{lightgray}\hline\arrayrulecolor{black}
            ~~\multirow{3}{*}{(d)} & COCO~\cite{COCO} & arXiv & 2016 & 43K & 39K \\ 
             & RCTW~\cite{RCTW} & ICDAR & 2017 & 65K & 8,186 \\
             & Uber~\cite{Uber} & CVPRW & 2017 & 285K & 92K \\
            \arrayrulecolor{lightgray}\hline\arrayrulecolor{black}
            ~~\multirow{4}{*}{(e)} & ArT~\cite{ArT} & ICDAR & 2019 & 50K & 29K \\
             & LSVT~\cite{LSVT} & ICDAR & 2019 & 383K & 34K \\
             & MLT19~\cite{MLT19} & ICDAR & 2019 & 89K & 46K \\
             & ReCTS~\cite{ReCTS} & ICDAR & 2019 & 147K & 23K \\
            \arrayrulecolor{lightgray}\hline\arrayrulecolor{black}
             & Total & $-$ & $-$ & 1.1M & 276K \\
            \midrule
           \multicolumn{4}{l}{\textbf{Real unlabeled datasets (Real-U)}} \\ 
             & Book32~\cite{book32} & arXiv & 2016 & 3.9M & 3.7M\\
             & TextVQA~\cite{TextVQA} & CVPR & 2019 &  551K & 463K\\
             & ST-VQA~\cite{ST-VQA} & ICCV & 2019 &  79K & 69K\\
            \arrayrulecolor{lightgray}\hline\arrayrulecolor{black}
             & Total & $-$ & $-$ & 4.6M & 4.2M\\
            \bottomrule
        \end{tabular}
        \end{adjustbox}
    \vspace{-2mm}
    \caption{Number of \textbf{training set} in public real datasets.}
    \label{tab:dataset}
    \end{center}
    \vspace{-8mm}
\end{table}

\subsection{Real Unlabeled Datasets}\label{sec:real_unlabel}
We consolidate three unlabeled datasets for semi- and self-supervised learning.
They contain scene images and do not have word region annotation. 
Thus, we use a pretrained text detector to crop words.
We use the detector~\cite{BDN-ReCTS}, which is not trained on synthetic data and won the ReCTS competition\footnote{https://rrc.cvc.uab.es/?ch=12\&com=evaluation\&task=3}.
Details are in the supplementary materials.

\PAR{Book32}~\cite{book32} is collected from Amazon Books, and consists of 208K book cover images in 32 categories. 
It contains many handwritten or curved texts.

\PAR{TextVQA}~\cite{TextVQA} is created for text-based visual question answering. 
It consists of 28K OpenImage V3~\cite{openimages} images from categories such as ``billboard'' and ``traffic sign.''

\PAR{ST-VQA}~\cite{ST-VQA} is created for scene text-based visual question answering.
It includes IC13, IC15, and COCO, and thus we excluded them.

\subsection{Preprocessing Real Datasets}\label{sec:preprocessing}
We conduct following processes before using real data:
\PAR{Excluding duplication between datasets}
Some well-known datasets (ICDAR03~(IC03)~\cite{IC03}, MLT17~\cite{MLT17}, and TotalText~\cite{totaltext}) are excluded because they are included in other datasets: IC13 inherits most of IC03, MLT19 includes MLT17, and ArT includes TotalText.
Also, CT and ArT have 122 duplicated word boxes, and we exclude them.

\PAR{Collecting only English words}
Some datasets are made for Chinese text recognition (RCTW, ArT, LSVT, ReCTS) or multilingual text recognition (MLT19).
Thus they contain languages other than English.
We only use words which consist of alphanumeric characters and symbols.

\PAR{Excluding don't care symbol}
Some texts have ``*'' or ``\#'', which denotes ``do not care about the text'' or ``characters hard to read.''
We exclude the texts containing them.

\PAR{Excluding vertical or $\pm$ 90 degree rotated texts}
Some datasets such as Uber-Text~\cite{Uber} contain many vertical texts or $\pm$ 90° rotated texts.
We mainly focus on horizontal texts and thus exclude vertical texts.
The images whose texts have more than two characters and whose height is greater than the width are excluded. 
For unlabeled data, the images whose height is greater than the width are excluded.

\PAR{Splitting training set to make validation set}
Most real datasets do not have validation set. 
Thus we split the training set of each dataset into training and validation sets.

In addition, we exclude texts longer than 25 characters following common practice~\cite{TRBA}.

\section{STR With Fewer Labels}\label{sec:method} 
In this section, we describe the STR models and semi- and self-supervised learning.
Although real data has increased as mentioned in \S\ref{sec:real_label_increment}, 
real data is still fewer than synthetic data at about 1.7\% of synthetic data.
To compensate for the low amount of data, we introduce a semi- and self-supervised learning framework to improve the STR with fewer labels. 
This is inspired by other computer vision tasks with fewer labels (high-fidelity image generation~\cite{high-fewerlabels} and ImageNet classification~\cite{s4l}).

\begin{figure}[t]
\includegraphics[width=0.9\linewidth, height=5.9cm]{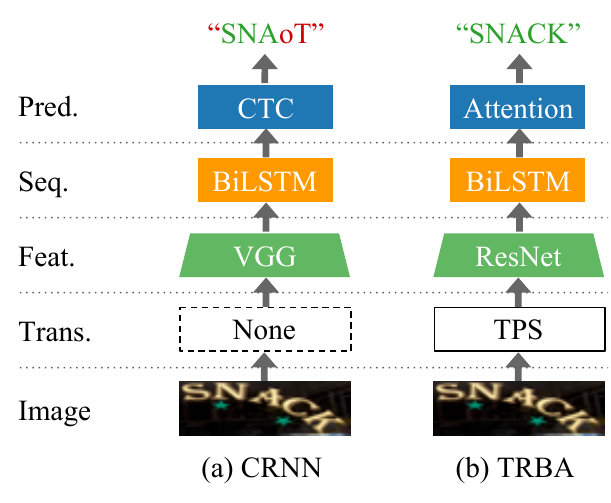}
    \centering
    \vspace{-4mm}
  \caption{Illustration of CRNN~\cite{CRNN} and TRBA~\cite{TRBA}.}
  \label{fig:model}
  \vspace{-1mm}
\end{figure}

\subsection{STR Model Framework}\label{sec:model}
According to \cite{TRBA}, STR is performed in four stages:

\begin{enumerate}
    \item \textbf{Transformation (Trans.)}: normalizes the perspective or curved text into a horizontal text. This is generally done by the Spatial Transformer Network~(STN)~\cite{STN}.
    \vspace{-2mm}
    
    \item \textbf{Feature extraction (Feat.)}: extracts visual feature representation from the input image. 
    This is generally performed by a module composed of convolutional neural networks (CNNs), such as VGG~\cite{VGG} and ResNet~\cite{ResNet}.
    \vspace{-2mm}
    
    \item \textbf{Sequence modeling (Seq.)}: 
    converts visual features to contextual features that capture context in the sequence of characters. 
    This is generally done by BiLSTM~\cite{BiLSTM}.
    
    \item \textbf{Prediction (Pred.)}: predicts the character sequence from contextual features. 
    This is generally done by a CTC~\cite{CTC} decoder or attention mechanism~\cite{NMT-attn}.
\end{enumerate}

For our experiments, we adopt two widely-used models from the STR benchmark~\cite{TRBA}:
CRNN\cite{CRNN} and TRBA\cite{TRBA}, as illustrated in Figure~\ref{fig:model}.
CRNN consists of None, VGG, BiLSTM, and CTC for each stage.
CRNN has lower accuracy than state-of-the-art methods, but CRNN is widely chosen for practical usage because it is fast and lightweight.
TRBA consists of a thin-plate spline~\cite{bookstein1989principal} transform-based STN (\textbf{T}PS), \textbf{R}esNet, \textbf{B}iLSTM, and \textbf{A}ttention for each stage.
As TRBA uses ResNet and attention mechanism, it is larger and slower than CRNN but has higher accuracy.

\begin{figure*}[t]
\centering
    \begin{subfigure}{0.49\linewidth} \centering
    \includegraphics[height=2.8cm]{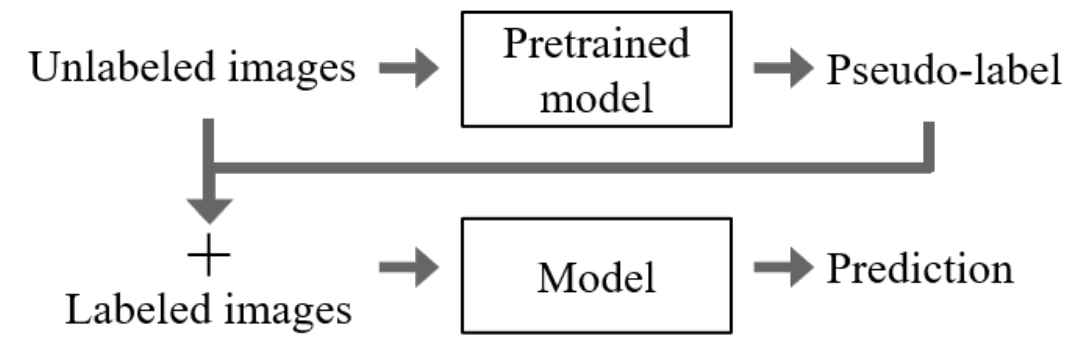}
     \vspace{-2mm}
     \caption{Pseudo-Label (PL)}\label{fig:PL}
    \end{subfigure}
    \begin{subfigure}{0.49\linewidth} \centering
    \includegraphics[width=0.99\linewidth, height=2.8cm]{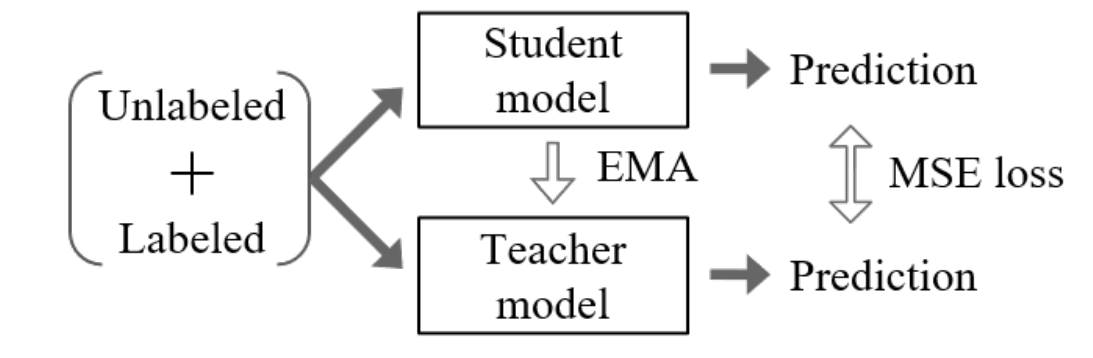}
     \vspace{-2mm}
     \caption{Mean Teacher (MT)}\label{fig:MT} 
    \end{subfigure}
\vspace{-2mm}
\caption{Illustration of Pseudo-Label~\cite{lee2013pseudo} and mean teacher~\cite{tarvainen2017mean}. $+$, EMA, and MSE denote union of labeled and unlabeled data, exponential moving average, and mean squared error, respectively.} 
\label{fig:PLMT}
\vspace{-2mm}
\end{figure*}

\subsection{Semi-Supervised Learning}\label{sec:semi}
Recently, various semi-supervised methods have been proposed and improved the performance with unlabeled data, particularly in image classification tasks~\cite{lee2013pseudo,tarvainen2017mean,VAT,s4l}.
Since large synthetic data is used for STR to compensate for the lack of data instead of using unlabeled data, studies on training STR with unlabeled data are rare. 
To the best of our knowledge, there is only one study that uses unlabeled data for the STR benchmark~\cite{PGT-RealSynthMix3}.
We introduce two simple yet effective semi-supervised methods for STR.

\PAR{Pseudo-Label~(PL)~\cite{lee2013pseudo}} is a simple approach that uses unlabeled data.
The process is as follows:
1) Train the model on labeled data.
2) Using the trained model, make predictions on unlabeled data and use them as pseudolabels.
3) Combine labeled and pseudolabeled data, and retrain the model on them.
Figure~\ref{fig:PL} illustrates PL.
Concurrent work~\cite{PGT-RealSynthMix3} also uses PL on the Book32 dataset. 
The researchers combine pseudolabeled and synthetic data, and use them as a training set.

\PAR{Mean Teacher~(MT)~\cite{tarvainen2017mean}} is a method that uses consistency regularization.
The process is as follows:
1) Prepare a model and a copy of the model.
2) Use the former as a student model and the latter as a teacher model.
3) Apply two random augmentations $\eta$ and $\eta'$ on the same mini-batch.
4) Input the former to the student model and the latter to the teacher model.
5) Calculate the mean squared error loss on their outputs.
6) Update the student model.
7) Update the teacher model with an exponential moving average (EMA) of the student model.
Figure~\ref{fig:MT} illustrates MT.

\subsection{Self-Supervised Learning}\label{sec:self}
Recently, self-supervised methods have shown promising results in computer vision tasks~\cite{RotNet,moco,SimCLR}.
Self-supervised learning is generally conducted in two stages: 1) Pretrain the model with a surrogate (pretext) task. 2) Using pretrained weights for initialization, train the model for the main task.
The pretext task is generally conducted on unlabeled data, and by learning a pretext task, the model obtains better feature maps for the main task.
In this study, we investigated two widely-used methods, RotNet~\cite{RotNet} and MoCo~\cite{moco}.

\PAR{RotNet~\cite{RotNet}} predicts the rotation of images as a pretext task.
The task is simple: rotate input images at 0, 90, 180, and 270 degrees, and the model recognizes the rotation applied to the image.

\PAR{Momentum Contrast~(MoCo)~\cite{moco}} is a contrastive learning method that can be applied to various pretext tasks.
Following \cite{moco}, we use an instance discrimination task~\cite{InstanceDiscriTask} as a pretext task.
The task consists of the following steps:
1) Prepare a model and a copy of the model.
2) Use the former as a query encoder and the latter as a momentum encoder.
3) Apply two random augmentations $\eta$ and $\eta'$ on the same mini-batch.
4) Input the former into a query encoder to make encoded queries $\bm{q}$.
5) Input the latter into a momentum encoder to make encoded keys $\bm{k}$.
6) Calculate the contrastive loss, called InfoNCE~\cite{InfoNCE}, on pairs of a query $q$ and a key $k$.
For a pair of $q$ and $k$, if they are derived from the same image, assign a positive, otherwise negative label.
7) Update the query encoder.
8) Update the momentum encoder with an moving average of the query encoder.

\section{Experiment and Analysis} \label{sec:experiments}
In this section, we present the results of our main experiments using STR with fewer real labels with a semi- and self-supervised learning framework.

\begin{table*}[t] 
    \begin{center}
        \begin{tabular}{@{}clclcccccc|c@{}}
            \toprule
            &&&& \multicolumn{7}{c}{Dataset name and \# of data} \\
            \cmidrule(l){5-11}
            & \multirow{2}{*}{Method} & \multirow{2}{*}{Year} 
            & \multirow{2}{*}{Train data} & IIIT & SVT & IC13 & IC15 & SP & \multicolumn{1}{c}{CT} & Total \\
            & &&& 3000 & 647 & 1015 & 2077 & 645 & \multicolumn{1}{c}{288} & 7672 \\
            \midrule
            \parbox[t]{2mm}{\multirow{9}{*}{\rotatebox[origin=c]{90}{\textbf{Reported results}}}}
& ASTER~\cite{ASTER} & 2018 & MJ+ST & 93.4 & 89.5 & 91.8 & 76.1 & 78.5 & 79.5 & 86.4\\
& ESIR~\cite{ESIR} & 2019 & MJ+ST & 93.3 & 90.2 & 91.3 & 76.9 & 79.6 & 83.3 & 86.8 \\
& MaskTextSpotter~\cite{Mask-textspotter} & 2019 & MJ+ST & \textbf{95.3} & 91.8 & \textbf{95.3} & 78.2 & 83.6 & 88.5 & 89.1 \\
& ScRN~\cite{ScRN} & 2019 & MJ+ST & 94.4 & 88.9 & 93.9 & 78.7 & 80.8 & 87.5 & 88.2 \\
& DAN~\cite{DAN} & 2020 & MJ+ST & 94.3 & 89.2 & 93.9 & 74.5 & 80.0 & 84.4 & 86.9 \\
& TextScanner~\cite{textscanner} & 2020 & MJ+ST & 93.9 & 90.1 & 92.9 & 79.4 & \textbf{84.3} & 83.3 & 88.3 \\
& SE-ASTER~\cite{SE-ASTER} & 2020 & MJ+ST & 93.8 & 89.6 & 92.8 & 80.0 & 81.4 & 83.6 & 88.2 \\
& RobustScanner~\cite{robustscanner} & 2020 & MJ+ST & \textbf{95.3} & 88.1 & 94.8 & 77.1 & 79.5 & \textbf{90.3} & 88.2 \\
& PlugNet~\cite{plugnet} & 2020 & MJ+ST & 94.4 & \textbf{92.3} & 95.0 & \textbf{82.2} & \textbf{84.3} & 85.0 & \textbf{89.8} \\
\midrule 
\midrule
\parbox[t]{2mm}{\multirow{8}{*}{\rotatebox[origin=c]{90}{\textbf{Our experiment}}}}
& CRNN-Original~\cite{CRNN}  & 2015  &  MJ  &78.2 & 80.8 & 86.7 & $-$ & $-$ & $-$ & $-$\\
& CRNN-Baseline-synth   &   & MJ+ST &84.3 & 78.9 & 88.8 & 61.5 & 64.8 & 61.3 & 75.8\\
& CRNN-Baseline-real     &  & Real-L  &83.5 & 75.5 & 86.3 & 62.2 & 60.9 & 64.7 & 74.8\\
& CRNN-PR &  & Real-L+U  &  89.8 & 84.3 & 90.9 & 73.1 & 74.6 & 82.3 & 83.4 \\
\cmidrule{2-11}
& TRBA-Original~\cite{TRBA} & 2019 & MJ+ST & 87.9 & 87.5 & 92.3 & 71.8 & 79.2 & 74.0 & 82.8 \\
& TRBA-Baseline-synth  &  & MJ+ST  &92.1 & 88.9 & 93.1 & 74.7 & 79.5 & 78.2 & 85.7 \\
& TRBA-Baseline-real &  & Real-L  & 93.5 & 87.5 & 92.6 & 76.0 & 78.7 & 86.1 & 86.6 \\
& TRBA-PR &  & Real-L+U & 94.8 & 91.3 & 94.0 & 80.6 & 82.7 & 88.1 & 89.3\\
\bottomrule
        \end{tabular}
    \vspace{-2mm}
    \caption{Accuracy of STR models on six benchmark datasets.
    We show the results reported in original papers.
    We present our results of CRNN and TRBA: Reproduced models (Baseline-synth), models trained only on real labels (Baseline-real), and our best setting (PR, combination of Pseudo-Label and RotNet).
    TRBA-PR has a competitive performance with state-of-the-art models.
    MJ, ST, Real-L, and Real-L+U denote MJSynth, SynthText, union of 11 real labeled datasets, and union of 11 real labeled and 3 unlabeled datasets in Table~\ref{tab:dataset}, respectively.
    In each column, top accuracy is shown in \textbf{bold}.
    }
    \label{tab:benchmark}
    \end{center}
    \vspace{-7mm}
\end{table*}

\subsection{Implementation Detail} \label{implementation}
We summarize our experimental settings.
More details of our settings are in our supplementary materials.

\PAR{Model and training strategy}
We use the code of the STR benchmark repository\footnote{https://github.com/clovaai/deep-text-recognition-benchmark}\cite{TRBA}, and use CRNN and TRBA as described in \S\ref{sec:model}.
We use the Adam~\cite{adam} optimizer and the one-cycle learning rate scheduler~\cite{super-convergence} with a maximum learning rate of 0.0005.
The number of iterations is 200K, and the batch size is 128.
As shown in Table~\ref{tab:dataset}, the number of training sets is imbalanced over the datasets.
To overcome the data imbalance, we sample the same number of data from each dataset to make a mini-batch.

\PAR{Dataset and model selection}
To train models on real data, we use the union of 11 real datasets (Real-L) listed in Table~\ref{tab:dataset}.
After preprocessing described in \S\ref{sec:preprocessing}, we have 276K training and 63K validation sets, and use them for training.
In all our experiments, we use the 63K validation set for model selection: select the model with the best accuracy on the validation set for evaluation.
We validate the model every 2,000 iterations, as in \cite{TRBA}.
To train models on synthetic data, we use the union of MJ (9M) and ST (7M).
For semi- and self-supervised learning, we use the union of 3 real unlabeled datasets (Real-U, 4.2M) listed in Table~\ref{tab:dataset}.

\PAR{Evaluation metric}
We use word-level accuracy on six benchmark datasets, as described in \S\ref{sec:benchmark_data}.
The accuracy is calculated only on the alphabet and digits, as done in \cite{ASTER}.
We calculate the total accuracy for comparison, which is the accuracy of the union of six benchmark datasets (7,672 in total).
In our study, accuracy indicates \textbf{total accuracy}.
For all experiments, we run three trials with different initializations and report averaged accuracies.

\subsection{Comparison to State-of-the-Art Methods}\label{exp:benchmark}
Table~\ref{tab:benchmark} lists the results of state-of-the-art methods and our experiments. 
For a fair comparison, we list the methods that use \textbf{only MJ and ST} for training, and evaluate six benchmarks: IIIT, SVT, IC13-1015, IC15-2077, SP, and CT.

Our reproduced models (Baseline-synth) has higher accuracies than in the original paper because we use different settings such as larger datasets (8M to 16M for CRNN and 14.4M to 16M for TRBA), different optimizer (Adam instead of AdaDelta~\cite{zeiler2012adadelta}), and learning rate scheduling.
Baseline-real is the model only trained on 11 real datasets.
CRNN-Baseline-real has an accuracy close to that of CRNN-Synth (74.8\% to 75.8\%), and TRBA-Baseline-real surpasses TRBA-Synth (86.6\% over 85.7\%).

\textit{TRBA with our best setting (TRBA-PR) trained on only real data has a competitive performance of 89.3\% with state-of-the-art methods.}
PR denotes the combination of Pseudo-Label and RotNet.
PR improves Baseline-real by +8.6\% for CRNN and +2.7\% for TRBA, and results in higher accuracy than Baseline-synth.
In the following sections, we analyze our best setting with ablation studies.

\subsection{Training Only on Real Labeled Data}\label{exp:labeled}
In this section, we present the results of training STR models only on real labeled data.

\PAR{Accuracy depending on dataset increment}
Table~\ref{tab:dataset} shows the increment of real data, and Figure~\ref{fig:data_increment} shows the accuracy improvement.
For 2019, when the number of the real training set is 276K, the accuracy of CRNN and TRBA trained only on real labeled data is close to that of synthetic data.
This indicates that we have enough real labeled data to train STR models satisfactorily, although the real labeled data is \textbf{only 1.7\%} of the synthetic data.

These results indicate that we need at least from 146K (Year 2017) to 276K (Year 2019) real data for training STR models.
However, according to \cite{TRBA}, the diversity of the training set can be more important than the number of training sets.
We use 11 datasets, which denotes high diversity, and thus we cannot simply conclude with ``276K is enough.''

To investigate the significance of real data, we also conduct an experiment that uses only 1.74\% of synthetic data (277K images), which is a similar amount to our real data.
This results in approximately 10\% lower accuracy than Baseline-real (65.1\% vs. 74.8\% for CRNN and 75.9\% vs. 86.6\% for TRBA).
This indicates that real data is far more significant than synthetic data.

\begin{table}[t] 
    \begin{center}
        \begin{tabular}{@{}llrr@{}}
            \toprule
            \multicolumn{2}{l}{Augmentation} & CRNN & TRBA \cr
            \midrule
            \multicolumn{2}{l}{Baseline-real} & 74.8 & 86.6 \\
            & + Blur & 75.7 & 86.8 \\
            & + Crop & 78.8 & 87.1 \\
            & + Rot  & 79.5 & 86.2 \\ 
            & + Blur + Crop  & 79.1  & \textbf{87.5} \\
            & + Blur + Rot  & 79.5  & 86.1  \\
            & + Crop + Rot  & \textbf{80.0} & 86.7  \\
            & + Blur + Crop  + Rot  & 78.9 & 86.6 \\
            \midrule
            \multicolumn{2}{l}{Baseline-synth} & 75.8 & 85.7 \\
            & + \textit{Aug.} & 73.4 & 85.2 \\
            \bottomrule
        \end{tabular}
    \vspace{-2mm}
    \caption{Improvement by simple data augmentations. \textit{Aug.} denotes the best augmentation setting in our experiments.}
    \label{tab:aug}
    \end{center}
    \vspace{-3mm}
\end{table} 

\PAR{Improvement by simple data augmentations}
Since our goal is to train an STR model with fewer labels,
to compensate for them, we find effective data augmentations.
Most STR methods do not use data augmentations~\cite{CRNN,ASTER,ESIR,TRBA,DAN,SE-ASTER,robustscanner,plugnet}.
We suppose that they do not use data augmentation because the synthetic data already includes augmented data.
In that case, if we apply further data augmentation above on already augmented data, then the results can be worse.
For example, applying a 45° rotation on the synthetic text, which was already rotated 45°, makes horizontal text to 90° rotated text. 
This can produce worse results.

However, if we use real data that do not contain data augmentations, then we can easily make improvements by data augmentation.
We investigate simple data augmentations.
Specifically, we use the Gaussian blur (Blur) to cope with blurry texts. 
We use a high ratio cropping (Crop), which slightly cuts the top, bottom, left, and right ends of the text, making STR models robust, and a rotation (Rot) for rotated, perspective, or curved texts.
The intensity of each augmentation affects the performance. 
We find the best intensities for them.
Table~\ref{tab:aug} shows the results of augmentations with best intensity and their combination.
The experiments with varying intensities of augmentations are in the supplementary materials.

Combinations of simple augmentations successfully improves the STR models.
For CRNN, the best setting (\textit{Aug.}) is the combination of Crop and Rot, which improves the accuracy by 5.2\% from Baseline-real.
For TRBA, the best setting (\textit{Aug.}) is the combination of Blur and Crop, which improves the accuracy by 0.9\% from Baseline-real.

We also apply the \textit{Aug.} to Baseline-synth, and the accuracy decreases.
We presume that the combination of already augmented data in synthetic data and \textit{Aug.} can be harmful to the performance.
For a similar case in our controlled experiments, the combination of Blur, Crop, and Rot has a lower accuracy than the combination of Crop and Rot.
These results indicate that the addition of augmentation can be harmful to the performance depending on the elements.

\subsection{Semi- and Self-Supervised Learning}\label{exp:semiself}
In addition to data augmentations, we further improve STR models by using unlabeled data (Real-U, listed in Table~\ref{tab:dataset}) with semi- and self-supervised methods, as described in \S\ref{sec:method}.
Table~\ref{tab:semiself} shows the results.

\begin{table}[t] 
    \begin{center}
        \begin{tabular}{@{}llll@{}}
            \toprule
            \multicolumn{2}{l}{Method} & CRNN & TRBA \cr
            \midrule
            \multicolumn{2}{l}{Baseline-real + \textit{Aug.}} & 80.0 & 87.5 \\
            & + PL                  & 82.8 \hl{(+2.8)}   & 89.2 \hl{(+1.7)} \\
            & + MT                  & 79.8 (-0.2)        & 87.1 (-0.4) \\
            \arrayrulecolor{lightgray}\midrule\arrayrulecolor{black}
            & + RotNet              & 81.3 \hl{(+1.3)}    & 87.5 \\
            & + MoCo                & 80.8 (+0.8)    & 86.7 (-0.8) \\
            \arrayrulecolor{lightgray}\midrule\arrayrulecolor{black}
            & + PL + RotNet         & \textbf{83.4 \hl{(+3.4)}} & \textbf{89.3 \hl{(+1.8)}} \\
            \bottomrule
        \end{tabular}
    \vspace{-1mm}
    \caption{Ablation study on semi- and self-supervised methods. 
    Gaps of at least 1.0 points are shown in \hl{green}.}
    \label{tab:semiself}
    \end{center}
    \vspace{-3mm}
\end{table}

\PAR{Pseudo-Label (PL) and Mean Teacher (MT):}
PL boosts the accuracy by 2.8\% for CRNN and 1.7\% for TRBA. 
MT decreases the accuracy by -0.2\% for CRNN and -0.4\% for TRBA.

\PAR{RotNet and MoCo:}
Following the common practice that pretrains CNN part of the model~\cite{RotNet,moco}, we pretrain VGG for CRNN and ResNet for TRBA.
We find that when we pretrain both TPS and ResNet in TRBA, the accuracy decreases sharply: -11.8\% with RotNet and -5.9\% with MoCo.
Thus, we only pretrain ResNet in TRBA.

For CRNN, RotNet and MoCo improve the accuracy by 1.3\% and 0.8\%, respectively. 
RotNet is slightly more effective than MoCo.
For TRBA, RotNet marginally improves the accuracy by +0.09\% (87.45\% to 87.54\%) and MoCo decreases the accuracy by -0.8\%.

\PAR{Combination of semi- and self-supervised methods}
The semi- and self-supervised methods in our experiments are independent, and thus we can combine them for further improvement.
We select PL and RotNet for the combination because they have better accuracy than MT and MoCo, respectively.
The PL method can be improved with a more accurate pretrained model to predict pseudolabels.
We use RotNet as the more accurate model.
Specifically, PL and RotNet are combined as follows:
1) Initialize the models with weights trained by the pretext task of RotNet.
2) Use RotNet to predict pseudolabels.
Table~\ref{tab:semiself} shows the results of the combination.

The combination of PL and RotNet has higher accuracy than solely using PL or RotNet.
This successfully improves the accuracy by +3.4\% from Baseline-real with \textit{Aug.} for CRNN and +1.8\% for TRBA.
Our best setting (PR) is the combination of \textit{Aug.}, PL, and RotNet. 

\begin{figure}[t]
\includegraphics[width=0.95\linewidth]{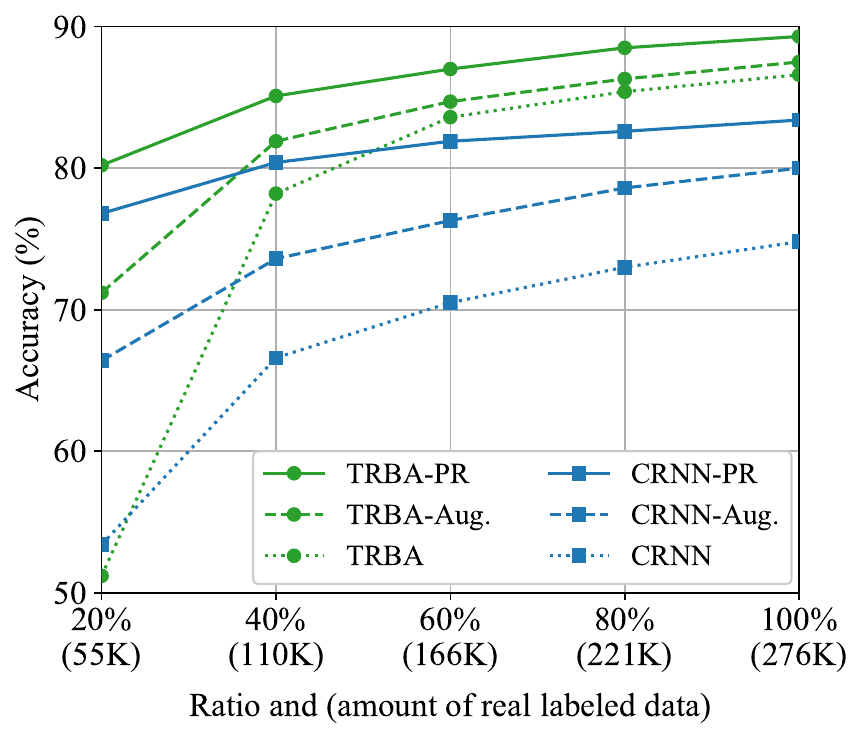}
    \centering
    \vspace{-3mm}
    \caption{Accuracy vs. amount of real labeled data.} 
  \label{fig:varying}
  \vspace{-3mm}
\end{figure}

\subsection{Varying Amount of Real Labeled Data} \label{sec:varying} 
Although Figure~\ref{fig:data_increment} shows the accuracy depending on dataset increment, the results are entangled with two factors: the amount of labeled data and the diversity of datasets.
We investigate the effect of the amount of labeled data by proportionally reducing each labeled dataset while maintaining the diversity of datasets (11 datasets).
The amount of unlabeled data is fixed.
Figure~\ref{fig:varying} shows the results.

Baseline-real is drastically dropped -13.2\% in accuracy for CRNN and -27.0\% for TRBA, with varying data ratios of 40\% to 20\%.
This shows that an accuracy cliff would appear here.
When our best augmentation setting (\textit{Aug.}) is applied, the accuracy improves fairly, especially with ratios of 20\%, by +13.0\% for CRNN and +20.0\% for TRBA.

PR with unlabeled data can substitute over 221K labeled data for CRNN and 110K labeled data for TRBA.
CRNN-PR with a ratio of 20\% exceeds Baseline-real with a ratio of 100\% by 2.0\%.
TRBA-PR with a ratio of 60\% exceeds Baseline-real with a ratio of 100\% by 0.4\%. 

The diversity of datasets can be more important than the amount of labeled data.
Comparing the Baseline with ratio 40\% (110K) to Year 2017 (146K) in Figure~\ref{fig:data_increment}, 
while the former has less data than the latter, the former has higher diversity than the latter (11 datasets vs. 7 datasets).
The former has higher accuracy than the latter: 66.6\% vs. 65.1\% for CRNN and 78.2\% vs. 75.1\% for TRBA.

\begin{table}[t] 
  \tabcolsep=0.10cm
    \begin{center}
        \begin{tabular}{@{}llllrr@{}}
            \toprule
            \multicolumn{3}{l}{Method} & Train Data & CRNN & TRBA \cr
            \midrule
            \multicolumn{3}{l}{\textbf{Fine-tuning}} \\
                && Baseline-synth & MJ+ST & 75.8 & 85.7 \\
                && + FT & Real-L & 82.1 & 90.0 \\
                && + FT w/PR  & Real-L+U & 76.6 & 87.5 \\
            \midrule
            \multicolumn{3}{l}{\textbf{From scratch}} \\
             && Baseline-real & Real-L          & 74.8 & 86.6 \\
             && Baseline-real & Real-L+MJ+ST    & 79.8 & 89.1 \\
             && PR & Real-L+U   & 83.4 & 89.3 \\
             && PR & Real-L+U+MJ+ST & 84.2 & 90.0\\
            \bottomrule
        \end{tabular}
    \vspace{-1mm}
    \caption{Training on both synthetic and real data.} 
    \label{tab:mix}
    \end{center}
    \vspace{-3mm}
\end{table}

\subsection{Training on Both Synthetic and Real Data}\label{sec:mix} 
In real scenarios, there is a case in which we have large synthetic data for the general domain and only fewer real data for the target domain.
We investigate if our best setting (PR) is also useful for this case by comparing other options.

\PAR{Fine-tuning on real data} 
Transfer learning with simple fine-tuning is a feasible option for such a case.
We conduct training STR models on large synthetic data (MJ and ST, 16M) and then fine-tuning on fewer real data (Real-L, 276K) for 40K iterations.

\PAR{Training from scratch}
Another option is training STR models on both of them from scratch.
We use the union of 11 real labeled and 2 synthetic datasets as a training set.

Table~\ref{tab:mix} shows the results.
Fine-tuning on real labeled data improves the accuracy by +6.3\% for CRNN and +4.3\% for TRBA.
Unexpectedly, fine-tuning with PR increases the accuracy (+0.8\% for CRNN and +1.8\% for TRBA) but has lower accuracy than fine-tuning only on real labeled data (76.6\% vs. 82.1\% for CRNN and 87.5\% vs. 90.0\% for TRBA).
This indicates that using semi- and self-supervised methods during fine-tuning can be harmful.

PR has higher accuracy than Baseline-real with synthetic data.
This shows that we can substitute the synthetic data with semi- and self-supervised methods that use unlabeled data.
For CRNN, PR with synthetic data has higher accuracy than the other settings.
This indicates that PR can be useful for training STR models when both large synthetic data and fewer real data are available.

\section{Conclusion} \label{sec:conclusion}
Since STR models have been trained on large synthetic data, training STR models on fewer real labels (STR with fewer labels) has not been sufficiently studied.
In this paper, we have focused on STR with fewer labels.
STR with fewer labels is considered difficult because there are only thousands of real data, resulting in low accuracy.
However, this is no longer the case.
We have shown that public real data has been accumulated over the years.
Although accumulated real data is only 1.7\% of the synthetic data, we can train STR models sufficiently by using it. 
We have further improved the performance by using simple data augmentations and introducing semi- and self-supervised methods with millions of real unlabeled data.
This work is a stepping stone toward STR with fewer labels, and we hope this work will facilitate future work on this topic.

{\small
\bibliographystyle{ieee_fullname}
\bibliography{main}

\begin{thebibliography}{10}\itemsep=-1pt

\bibitem{TRBA}
Jeonghun Baek, Geewook Kim, Junyeop Lee, Sungrae Park, Dongyoon Han, Sangdoo
  Yun, Seong~Joon Oh, and Hwalsuk Lee.
\newblock What is wrong with scene text recognition model comparisons? dataset
  and model analysis.
\newblock In {\em ICCV}, 2019.

\bibitem{NMT-attn}
Dzmitry Bahdanau, Kyunghyun Cho, and Yoshua Bengio.
\newblock Neural machine translation by jointly learning to align and
  translate.
\newblock In {\em ICLR}, 2015.

\bibitem{Bhunia_2017}
Ayan~Kumar Bhunia, Gautam Kumar, Partha~Pratim Roy, R. Balasubramanian, and
  Umapada Pal.
\newblock Text recognition in scene image and video frame using color channel
  selection.
\newblock {\em Multimedia Tools and Applications}, 2017.

\bibitem{ST-VQA}
Ali~Furkan Biten, Ruben Tito, Andres Mafla, Lluis Gomez, Mar{\c{c}}al Rusinol,
  Ernest Valveny, CV Jawahar, and Dimosthenis Karatzas.
\newblock Scene text visual question answering.
\newblock In {\em ICCV}, 2019.

\bibitem{bookstein1989principal}
Fred~L. Bookstein.
\newblock Principal warps: Thin-plate splines and the decomposition of
  deformations.
\newblock {\em TPAMI}, 1989.

\bibitem{SimCLR}
Ting Chen, Simon Kornblith, Mohammad Norouzi, and Geoffrey Hinton.
\newblock A simple framework for contrastive learning of visual
  representations.
\newblock In {\em ICML}, 2020.

\bibitem{ArT}
Chee~Kheng Chng, Yuliang Liu, Yipeng Sun, Chun~Chet Ng, Canjie Luo, Zihan Ni,
  ChuanMing Fang, Shuaitao Zhang, Junyu Han, Errui Ding, et~al.
\newblock Icdar2019 robust reading challenge on arbitrary-shaped text-rrc-art.
\newblock In {\em ICDAR}, 2019.

\bibitem{totaltext}
Chee~Kheng Ch’ng, Chee~Seng Chan, and Chenglin Liu.
\newblock Total-text: Towards orientation robustness in scene text detection.
\newblock {\em IJDAR}, 2020.

\bibitem{RotNet}
Spyros Gidaris, Praveer Singh, and Nikos Komodakis.
\newblock Unsupervised representation learning by predicting image rotations.
\newblock In {\em ICLR}, 2018.

\bibitem{CTC}
Alex Graves, Santiago Fern{\'a}ndez, Faustino Gomez, and J{\"u}rgen
  Schmidhuber.
\newblock Connectionist temporal classification: labelling unsegmented sequence
  data with recurrent neural networks.
\newblock In {\em ICML}, 2006.

\bibitem{BiLSTM}
Alex Graves, Marcus Liwicki, Santiago Fern{\'a}ndez, Roman Bertolami, Horst
  Bunke, and J{\"u}rgen Schmidhuber.
\newblock A novel connectionist system for unconstrained handwriting
  recognition.
\newblock {\em TPAMI}, 2009.

\bibitem{SynthText}
Ankush Gupta, Andrea Vedaldi, and Andrew Zisserman.
\newblock Synthetic data for text localisation in natural images.
\newblock In {\em CVPR}, 2016.

\bibitem{moco}
Kaiming He, Haoqi Fan, Yuxin Wu, Saining Xie, and Ross Girshick.
\newblock Momentum contrast for unsupervised visual representation learning.
\newblock In {\em CVPR}, 2020.

\bibitem{he2015delving}
Kaiming He, Xiangyu Zhang, Shaoqing Ren, and Jian Sun.
\newblock Delving deep into rectifiers: Surpassing human-level performance on
  imagenet classification.
\newblock In {\em ICCV}, 2015.

\bibitem{ResNet}
Kaiming He, Xiangyu Zhang, Shaoqing Ren, and Jian Sun.
\newblock Deep residual learning for image recognition.
\newblock In {\em CVPR}, 2016.

\bibitem{book32}
Brian~Kenji Iwana, Syed Tahseen~Raza Rizvi, Sheraz Ahmed, Andreas Dengel, and
  Seiichi Uchida.
\newblock Judging a book by its cover.
\newblock {\em arXiv:1610.09204}, 2016.

\bibitem{MJSynth}
Max Jaderberg, Karen Simonyan, Andrea Vedaldi, and Andrew Zisserman.
\newblock Synthetic data and artificial neural networks for natural scene text
  recognition.
\newblock In {\em Workshop on Deep Learning, NeurIPS}, 2014.

\bibitem{STN}
Max Jaderberg, Karen Simonyan, Andrew Zisserman, et~al.
\newblock Spatial transformer networks.
\newblock In {\em NeurIPS}, 2015.

\bibitem{PGT-RealSynthMix3}
Kl{\'a}ra Janou{\v{s}}kov{\'a}, Jiri Matas, Lluis Gomez, and Dimosthenis
  Karatzas.
\newblock Text recognition--real world data and where to find them.
\newblock {\em arXiv:2007.03098}, 2020.

\bibitem{IC15}
Dimosthenis Karatzas, Lluis Gomez-Bigorda, Anguelos Nicolaou, Suman Ghosh,
  Andrew Bagdanov, Masakazu Iwamura, Jiri Matas, Lukas Neumann,
  Vijay~Ramaseshan Chandrasekhar, Shijian Lu, et~al.
\newblock Icdar 2015 competition on robust reading.
\newblock In {\em ICDAR}, 2015.

\bibitem{IC13}
Dimosthenis Karatzas, Faisal Shafait, Seiichi Uchida, Masakazu Iwamura,
  Lluis~Gomez i Bigorda, Sergi~Robles Mestre, Joan Mas, David~Fernandez Mota,
  Jon~Almazan Almazan, and Lluis~Pere De~Las~Heras.
\newblock Icdar 2013 robust reading competition.
\newblock In {\em ICDAR}, 2013.

\bibitem{adam}
Diederik~P. Kingma and Jimmy Ba.
\newblock Adam: A method for stochastic optimization.
\newblock In {\em ICLR}, 2015.

\bibitem{openimages}
Ivan Krasin, Tom Duerig, Neil Alldrin, Vittorio Ferrari, Sami Abu-El-Haija,
  Alina Kuznetsova, Hassan Rom, Jasper Uijlings, Stefan Popov, Andreas Veit,
  Serge Belongie, Victor Gomes, Abhinav Gupta, Chen Sun, Gal Chechik, David
  Cai, Zheyun Feng, Dhyanesh Narayanan, and Kevin Murphy.
\newblock Openimages: A public dataset for large-scale multi-label and
  multi-class image classification.
\newblock {\em Dataset available from https://github.com/openimages}, 2017.

\bibitem{lee2013pseudo}
Dong-Hyun Lee.
\newblock Pseudo-label: The simple and efficient semi-supervised learning
  method for deep neural networks.
\newblock In {\em Workshop on challenges in representation learning, ICML},
  2013.

\bibitem{SAR}
Hui Li, Peng Wang, Chunhua Shen, and Guyu Zhang.
\newblock Show, attend and read: A simple and strong baseline for irregular
  text recognition.
\newblock In {\em AAAI}, 2019.

\bibitem{Mask-textspotter}
Minghui Liao, Pengyuan Lyu, Minghang He, Cong Yao, Wenhao Wu, and Xiang Bai.
\newblock Mask textspotter: An end-to-end trainable neural network for spotting
  text with arbitrary shapes.
\newblock {\em TPAMI}, 2019.

\bibitem{Textbox++2018TIP-CRNNusedforSTS}
Minghui Liao, Baoguang Shi, and Xiang Bai.
\newblock Textboxes++: A single-shot oriented scene text detector.
\newblock {\em TIP}, 2018.

\bibitem{TextboxAAAI17-CRNNusedforSTS}
Minghui Liao, Baoguang Shi, Xiang Bai, Xinggang Wang, and Wenyu Liu.
\newblock Textboxes: A fast text detector with a single deep neural network.
\newblock In {\em AAAI}, 2017.

\bibitem{MSCOCO}
Tsung-Yi Lin, Michael Maire, Serge Belongie, James Hays, Pietro Perona, Deva
  Ramanan, Piotr Doll{\'a}r, and C~Lawrence Zitnick.
\newblock Microsoft coco: Common objects in context.
\newblock In {\em ECCV}, 2014.

\bibitem{scatter}
Ron Litman, Oron Anschel, Shahar Tsiper, Roee Litman, Shai Mazor, and R
  Manmatha.
\newblock Scatter: selective context attentional scene text recognizer.
\newblock In {\em CVPR}, 2020.

\bibitem{liu2020abcnet-CRNNusedforSTS}
Yuliang Liu, Hao Chen, Chunhua Shen, Tong He, Lianwen Jin, and Liangwei Wang.
\newblock Abcnet: Real-time scene text spotting with adaptive bezier-curve
  network.
\newblock In {\em CVPR}, 2020.

\bibitem{BDN-ReCTS}
Yuliang Liu, Tong He, Hao Chen, Xinyu Wang, Canjie Luo, Shuaitao Zhang, Chunhua
  Shen, and Lianwen Jin.
\newblock Exploring the capacity of sequential-free box discretization network
  for omnidirectional scene text detection.
\newblock {\em arXiv:1912.09629}, 2019.

\bibitem{CTW1500}
Yuliang Liu, Lianwen Jin, Shuaitao Zhang, Canjie Luo, and Sheng Zhang.
\newblock Curved scene text detection via transverse and longitudinal sequence
  connection.
\newblock {\em Pattern Recognition}, 2019.

\bibitem{BDN}
Yuliang Liu, Sheng Zhang, Lianwen Jin, Lele Xie, Yaqiang Wu, and Zhepeng Wang.
\newblock Omnidirectional scene text detection with sequential-free box
  discretization.
\newblock In {\em IJCAI}, 2019.

\bibitem{long2020unrealtext}
Shangbang Long and Cong Yao.
\newblock Unrealtext: Synthesizing realistic scene text images from the unreal
  world.
\newblock {\em arXiv:2003.10608}, 2020.

\bibitem{NIPS2016_shape}
Xinghua Lou, Ken Kansky, Wolfgang Lehrach, CC Laan, Bhaskara Marthi, D.
  Phoenix, and Dileep George.
\newblock Generative shape models: Joint text recognition and segmentation with
  very little training data.
\newblock In {\em NeurIPS}, 2016.

\bibitem{IC03}
Simon~M Lucas, Alex Panaretos, Luis Sosa, Anthony Tang, Shirley Wong, and
  Robert Young.
\newblock Icdar 2003 robust reading competitions.
\newblock In {\em ICDAR}, 2003.

\bibitem{high-fewerlabels}
Mario Lucic, Michael Tschannen, Marvin Ritter, Xiaohua Zhai, Olivier Bachem,
  and Sylvain Gelly.
\newblock High-fidelity image generation with fewer labels.
\newblock In {\em ICML}, 2019.

\bibitem{IIIT5K}
Anand Mishra, Karteek Alahari, and CV Jawahar.
\newblock Scene text recognition using higher order language priors.
\newblock In {\em BMVC}, 2012.

\bibitem{mishra2016enhancing}
Anand Mishra, Karteek Alahari, and CV Jawahar.
\newblock Enhancing energy minimization framework for scene text recognition
  with top-down cues.
\newblock {\em Computer Vision and Image Understanding}, 2016.

\bibitem{VAT}
Takeru Miyato, Shin-ichi Maeda, Masanori Koyama, and Shin Ishii.
\newblock Virtual adversarial training: a regularization method for supervised
  and semi-supervised learning.
\newblock {\em TPAMI}, 2018.

\bibitem{plugnet}
Yongqiang Mou, Lei Tan, Hui Yang, Jingying Chen, Leyuan Liu, Rui Yan, and
  Yaohong Huang.
\newblock Plugnet: Degradation aware scene text recognition supervised by a
  pluggable super-resolution unit.
\newblock In {\em ECCV}, 2020.

\bibitem{MLT19}
Nibal Nayef, Yash Patel, Michal Busta, Pinaki~Nath Chowdhury, Dimosthenis
  Karatzas, Wafa Khlif, Jiri Matas, Umapada Pal, Jean-Christophe Burie,
  Cheng-lin Liu, et~al.
\newblock Icdar2019 robust reading challenge on multi-lingual scene text
  detection and recognition—rrc-mlt-2019.
\newblock In {\em ICDAR}, 2019.

\bibitem{MLT17}
Nibal Nayef, Fei Yin, Imen Bizid, Hyunsoo Choi, Yuan Feng, Dimosthenis
  Karatzas, Zhenbo Luo, Umapada Pal, Christophe Rigaud, Joseph Chazalon, et~al.
\newblock Icdar2017 robust reading challenge on multi-lingual scene text
  detection and script identification-rrc-mlt.
\newblock In {\em ICDAR}, 2017.

\bibitem{InfoNCE}
Aaron van~den Oord, Yazhe Li, and Oriol Vinyals.
\newblock Representation learning with contrastive predictive coding.
\newblock {\em arXiv:1807.03748}, 2018.

\bibitem{PyTorch}
Adam Paszke, Sam Gross, Francisco Massa, Adam Lerer, James Bradbury, Gregory
  Chanan, Trevor Killeen, Zeming Lin, Natalia Gimelshein, Luca Antiga, Alban
  Desmaison, Andreas Kopf, Edward Yang, Zachary DeVito, Martin Raison, Alykhan
  Tejani, Sasank Chilamkurthy, Benoit Steiner, Lu Fang, Junjie Bai, and Soumith
  Chintala.
\newblock Pytorch: An imperative style, high-performance deep learning library.
\newblock In {\em NeurIPS}, 2019.

\bibitem{patel2020learning-ECCV-TRBAused}
Yash Patel, Tomas Hodan, and Jiri Matas.
\newblock Learning surrogates via deep embedding.
\newblock In {\em ECCV}, 2020.

\bibitem{SVTP}
Trung~Quy Phan, Palaiahnakote Shivakumara, Shangxuan Tian, and Chew~Lim Tan.
\newblock Recognizing text with perspective distortion in natural scenes.
\newblock In {\em ICCV}, 2013.

\bibitem{SE-ASTER}
Zhi Qiao, Yu Zhou, Dongbao Yang, Yucan Zhou, and Weiping Wang.
\newblock Seed: Semantics enhanced encoder-decoder framework for scene text
  recognition.
\newblock In {\em CVPR}, 2020.

\bibitem{CUTE80}
Anhar Risnumawan, Palaiahankote Shivakumara, Chee~Seng Chan, and Chew~Lim Tan.
\newblock A robust arbitrary text detection system for natural scene images.
\newblock {\em ESWA}, 2014.

\bibitem{CRNN}
Baoguang Shi, Xiang Bai, and Cong Yao.
\newblock An end-to-end trainable neural network for image-based sequence
  recognition and its application to scene text recognition.
\newblock {\em TPAMI}, 2016.

\bibitem{RARE}
Baoguang Shi, Xinggang Wang, Pengyuan Lyu, Cong Yao, and Xiang Bai.
\newblock Robust scene text recognition with automatic rectification.
\newblock In {\em CVPR}, 2016.

\bibitem{ASTER}
Baoguang Shi, Mingkun Yang, Xinggang Wang, Pengyuan Lyu, Cong Yao, and Xiang
  Bai.
\newblock Aster: An attentional scene text recognizer with flexible
  rectification.
\newblock {\em TPAMI}, 2018.

\bibitem{RCTW}
Baoguang Shi, Cong Yao, Minghui Liao, Mingkun Yang, Pei Xu, Linyan Cui, Serge
  Belongie, Shijian Lu, and Xiang Bai.
\newblock Icdar2017 competition on reading chinese text in the wild (rctw-17).
\newblock In {\em ICDAR}, 2017.

\bibitem{VGG}
Karen Simonyan and Andrew Zisserman.
\newblock Very deep convolutional networks for large-scale image recognition.
\newblock In {\em ICLR}, 2015.

\bibitem{TextVQA}
Amanpreet Singh, Vivek Natarajan, Meet Shah, Yu Jiang, Xinlei Chen, Dhruv
  Batra, Devi Parikh, and Marcus Rohrbach.
\newblock Towards vqa models that can read.
\newblock In {\em CVPR}, 2019.

\bibitem{super-convergence}
Leslie~N. Smith and Nicholay Topin.
\newblock {Super-convergence: very fast training of neural networks using large
  learning rates}.
\newblock {\em AI/ML for MDO}, 2019.

\bibitem{srivastava2014dropout}
Nitish Srivastava, Geoffrey Hinton, Alex Krizhevsky, Ilya Sutskever, and Ruslan
  Salakhutdinov.
\newblock Dropout: a simple way to prevent neural networks from overfitting.
\newblock {\em JMLR}, 2014.

\bibitem{sun2019LSVT}
Yipeng Sun, Jiaming Liu, Wei Liu, Junyu Han, Errui Ding, and Jingtuo Liu.
\newblock Chinese street view text: Large-scale chinese text reading with
  partially supervised learning.
\newblock In {\em ICCV}, 2019.

\bibitem{LSVT}
Yipeng Sun, Zihan Ni, Chee-Kheng Chng, Yuliang Liu, Canjie Luo, Chun~Chet Ng,
  Junyu Han, Errui Ding, Jingtuo Liu, Dimosthenis Karatzas, et~al.
\newblock Icdar 2019 competition on large-scale street view text with partial
  labeling-rrc-lsvt.
\newblock In {\em ICDAR}, 2019.

\bibitem{tarvainen2017mean}
Antti Tarvainen and Harri Valpola.
\newblock Mean teachers are better role models: Weight-averaged consistency
  targets improve semi-supervised deep learning results.
\newblock In {\em NeurIPS}, 2017.

\bibitem{COCO}
Andreas Veit, Tomas Matera, Lukas Neumann, Jiri Matas, and Serge Belongie.
\newblock Coco-text: Dataset and benchmark for text detection and recognition
  in natural images.
\newblock {\em arXiv:1601.07140}, 2016.

\bibitem{textscanner}
Zhaoyi Wan, Mingling He, Haoran Chen, Xiang Bai, and Cong Yao.
\newblock Textscanner: Reading characters in order for robust scene text
  recognition.
\newblock In {\em AAAI}, 2020.

\bibitem{SVT}
Kai Wang, Boris Babenko, and Serge Belongie.
\newblock End-to-end scene text recognition.
\newblock In {\em ICCV}, 2011.

\bibitem{DAN}
Tianwei Wang, Yuanzhi Zhu, Lianwen Jin, Canjie Luo, Xiaoxue Chen, Yaqiang Wu,
  Qianying Wang, and Mingxiang Cai.
\newblock Decoupled attention network for text recognition.
\newblock In {\em AAAI}, 2020.

\bibitem{wang2020scene-SR-ECCV}
Wenjia Wang, Enze Xie, Xuebo Liu, Wenhai Wang, Ding Liang, Chunhua Shen, and
  Xiang Bai.
\newblock Scene text image super-resolution in the wild.
\newblock In {\em ECCV}, 2020.

\bibitem{InstanceDiscriTask}
Zhirong Wu, Yuanjun Xiong, Stella~X Yu, and Dahua Lin.
\newblock Unsupervised feature learning via non-parametric instance
  discrimination.
\newblock In {\em CVPR}, 2018.

\bibitem{Xu_2020_CVPR-WhatMachines-TRBAused}
Xing Xu, Jiefu Chen, Jinhui Xiao, Lianli Gao, Fumin Shen, and Heng~Tao Shen.
\newblock What machines see is not what they get: Fooling scene text
  recognition models with adversarial text images.
\newblock In {\em CVPR}, 2020.

\bibitem{xu2020_ACMMM-adversarial-TRBAused}
Xing Xu, Jiefu Chen, Jinhui Xiao, Zheng Wang, Yang Yang, and Heng~Tao Shen.
\newblock Learning optimization-based adversarial perturbations for attacking
  sequential recognition models.
\newblock In {\em ACMMM}, 2020.

\bibitem{ScRN}
Mingkun Yang, Yushuo Guan, Minghui Liao, Xin He, Kaigui Bian, Song Bai, Cong
  Yao, and Xiang Bai.
\newblock Symmetry-constrained rectification network for scene text
  recognition.
\newblock In {\em ICCV}, 2019.

\bibitem{SRN}
Deli Yu, Xuan Li, Chengquan Zhang, Tao Liu, Junyu Han, Jingtuo Liu, and Errui
  Ding.
\newblock Towards accurate scene text recognition with semantic reasoning
  networks.
\newblock In {\em CVPR}, 2020.

\bibitem{robustscanner}
Xiaoyu Yue, Zhanghui Kuang, Chenhao Lin, Hongbin Sun, and Wayne Zhang.
\newblock Robustscanner: Dynamically enhancing positional clues for robust text
  recognition.
\newblock In {\em ECCV}, 2020.

\bibitem{zeiler2012adadelta}
Matthew~D Zeiler.
\newblock Adadelta: an adaptive learning rate method.
\newblock {\em arXiv:1212.5701}, 2012.

\bibitem{s4l}
Xiaohua Zhai, Avital Oliver, Alexander Kolesnikov, and Lucas Beyer.
\newblock S4l: Self-supervised semi-supervised learning.
\newblock In {\em ICCV}, 2019.

\bibitem{ESIR}
Fangneng Zhan and Shijian Lu.
\newblock Esir: End-to-end scene text recognition via iterative image
  rectification.
\newblock In {\em CVPR}, 2019.

\bibitem{ArT-won-onlyReal}
Jinjin Zhang, Wei Wang, Di Huang, Qingjie Liu, and Yunhong Wang.
\newblock A feasible framework for arbitrary-shaped scene text recognition.
\newblock {\em arXiv:1912.04561}, 2019.

\bibitem{ReCTS}
Rui Zhang, Yongsheng Zhou, Qianyi Jiang, Qi Song, Nan Li, Kai Zhou, Lei Wang,
  Dong Wang, Minghui Liao, Mingkun Yang, et~al.
\newblock Icdar 2019 robust reading challenge on reading chinese text on
  signboard.
\newblock In {\em ICDAR}, 2019.

\bibitem{Uber}
Ying Zhang, Lionel Gueguen, Ilya Zharkov, Peter Zhang, Keith Seifert, and Ben
  Kadlec.
\newblock Uber-text: A large-scale dataset for optical character recognition
  from street-level imagery.
\newblock In {\em Scene Understanding Workshop, CVPR}, 2017.

\end{thebibliography}
}

\clearpage
\appendix
\section*{Supplementary Material}
\section{Contents}\label{sup:contents}

\noindent Supplement~\ref{sup:necess} : \textbf{Needs for STR with Fewer Real Labels} 
    \begin{itemize}
        \vspace{-1mm}
        \item We complement the needs for STR with fewer real labels by illustrating the detailed examples.
    \end{itemize}
        
\noindent Supplement~\ref{sup:data} : \textbf{STR Datasets - Details and More Examples} 
    \begin{itemize}
    \vspace{-1mm}
        \item We describe the detail of preprocessing in \S\ref{sec:consoli} and show more examples of public real data.
    \end{itemize}
    
\noindent Supplement~\ref{sup:method} : \textbf{STR With Fewer Labels - Details}
    \begin{itemize}
    \vspace{-1mm}
        \item We describe the details of STR models and semi- and self-supervised methods in \S\ref{sec:method}.
    \end{itemize}

\noindent Supplement~\ref{sup:experiment} : \textbf{Experiment and Analysis - Details}
    \begin{itemize}
    \vspace{-1mm}
        \item We provide the details and comprehensive results of our experiments in \S\ref{sec:experiments}.
    \end{itemize}

\section{Needs for STR with Fewer Real Labels}\label{sup:necess}
In this section, we complement the necessity of training STR models only on fewer real labels (STR with fewer labels).
The study of STR with fewer labels generally \textit{aims to exploit a few real labels efficiently}.
We describe the detailed examples of two needs as mentioned in \S\ref{sec:introduction}, and introduce an additional need based on our experimental results.

\begin{figure}[t]
\centering
    \begin{subfigure}{0.98\linewidth} \centering
     \includegraphics[width=0.485\linewidth, height=2.5cm]{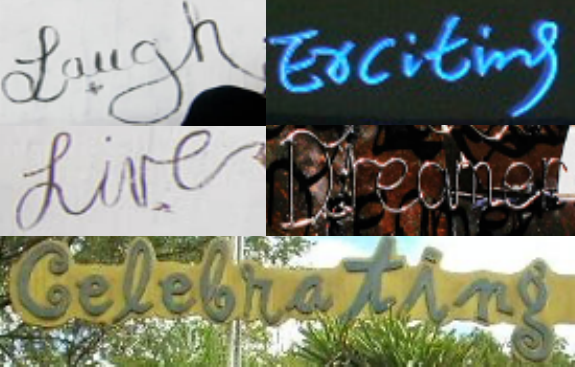}
     \includegraphics[width=0.485\linewidth, height=2.5cm]{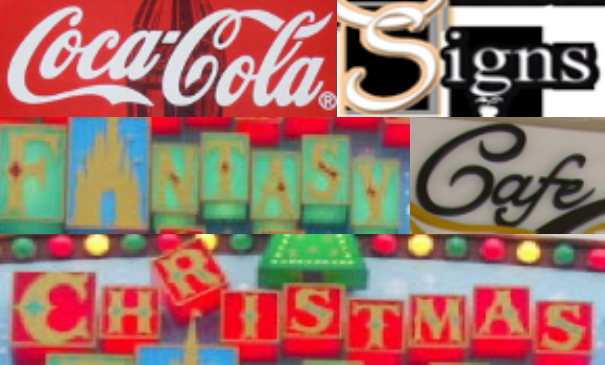}
     \vspace{-1mm}
     \caption{Examples from benchmark datasets for evaluation.}
    \end{subfigure}
    
    \vspace{1mm}
    \begin{subfigure}{0.98\linewidth} \centering
     \includegraphics[width=0.485\linewidth, height=2.5cm]{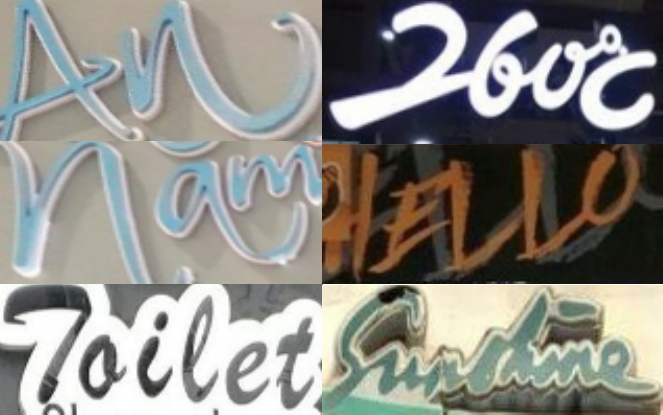}
     \includegraphics[width=0.485\linewidth, height=2.5cm]{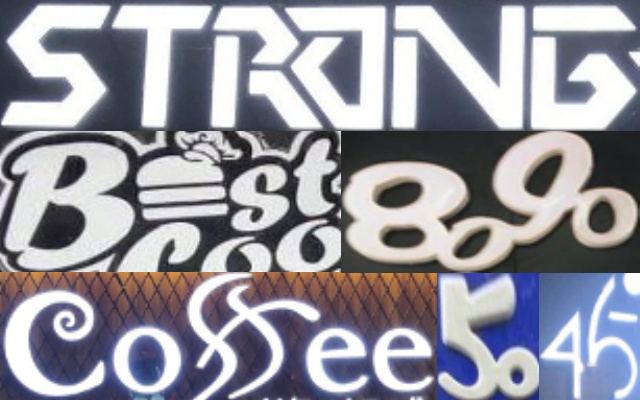}
     \vspace{-1mm}
     \caption{Examples from real datasets other than benchmark datasets.}
    \end{subfigure}
\vspace{-2mm}
\caption{Handwritten or artistic texts in public real data.}
\label{sup-fig:handwritten}
\vspace{-2mm}
\end{figure}

\PAR{Need \#1: To recognize handwritten or artistic data in public real datasets}
Figure~\ref{sup-fig:handwritten} shows the handwritten or artistic data in public real datasets.
It is difficult to generate them with the current synthetic engine~\cite{MJSynth,SynthText}.
If we have appropriate handwritten fonts, we can generate similar texts with handwritten texts synthetically and cover some of the handwritten texts.
However, because the number of fonts (about thousands) is quite lower than the number of people, there can still be uncovered handwritten texts by handwritten fonts.
Furthermore, generating artistic texts with handwritten fonts is difficult: artistic texts such as text logo and calligraphy as shown in Figure~\ref{sup-fig:handwritten}.

\begin{figure}[t]
\begin{center}
\includegraphics[width=\linewidth]{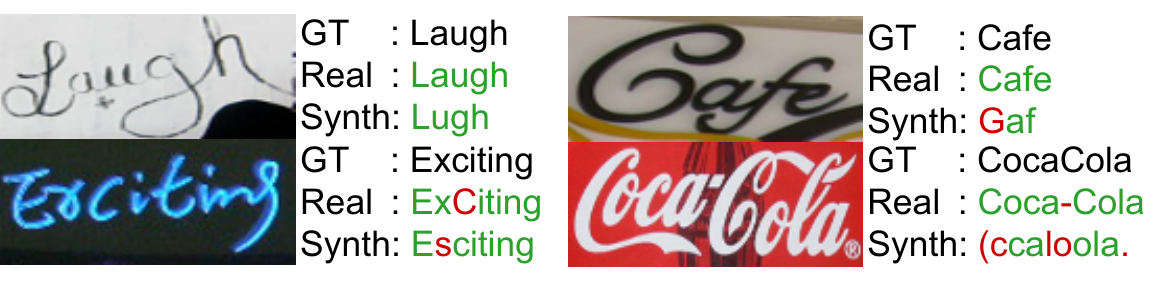}
\end{center}
\vspace{-6mm}
\caption{Predictions on handwritten or artistic texts. 
GT, Real, and Synth denote the ground truth, the prediction of TRBA-Baseline-real, and TRBA-Baseline-synth, respectively.}
\vspace{-3mm}
\label{sup-fig:qualitative}
\end{figure}

In this case, exploiting the few real data of them can be more efficient rather than generating synthetic data close to them.
Namely, we need STR with fewer labels in this case.
Figure~\ref{sup-fig:qualitative} shows the predictions by trained models (TRBA-Baseline-real and -synth) in our experiments. 
The model trained with real data (Real) has better performance than that with synthetic data (Synth).
These results show that exploiting real data can be useful for these types of texts.

\begin{table*}[t] 
  \tabcolsep=0.13cm
    \begin{center}
        \begin{adjustbox}{width=0.975\linewidth}
        \begin{tabular}{@{}l|ccccccc|c|ccccccc|c@{}}
            \toprule
            & \multicolumn{8}{c|}{Model: CRNN} & \multicolumn{8}{c}{Model: TRBA} \\ 
            Method & COCO & RCTW & Uber & ArT & LSVT & MLT19 & \multicolumn{1}{c}{ReCTS} & Avg. & COCO & RCTW & Uber & ArT & LSVT & MLT19 & \multicolumn{1}{c}{ReCTS} & Avg.\\
            \midrule
            Baseline-synth & 37.4 & 50.3 & 32.1 & 48.1 & 50.3 & 74.6 & 73.6 & 52.3 & 50.2 & 59.1 & 36.7 & 57.6 & 58.0 & 80.3 & 80.6 & 60.4 \\
            Baseline-real & 46.3 & 54.3 & 45.8 & 48.1 & 58.6 & 78.2 & 74.7 & 58.0 & 62.7 & 67.7 & 52.7 & 63.2 & 68.7 & 85.8 & 83.4 & 69.2 \\
            PR & 56.6 & 61.1 & 44.8 & 58.7 & 62.2 & 82.5 & 80.9 & 63.8 & 66.9 & 71.5 & 54.2 & 66.7 & 73.5 & 87.8 & 85.6 & 72.3 \\
            \bottomrule
        \end{tabular}
        \end{adjustbox}
    \vspace{-1mm}
    \caption{Accuracy of seven evaluation datasets: COCO, RCTW, Uber, ArT, LSVT, MLT19, and ReCTS.
    Avg. denotes the averaged accuracies of seven datasets.
    The number of evaluation sets of each datasets is described in Table~\ref{sup-tab:split}.
    }
    \label{sup-tab:eval-seven-addition}
    \end{center}
    \vspace{-3mm}
\end{table*}

\PAR{Need \#2: To recognize language other than English (LOTE) without synthetic data}
In the other case, when we have to recognize LOTE, there are not always synthetic data for LOTE.
For such a case, we should recognize LOTE without synthetic data or generate synthetic data.

However, generating appropriate synthetic data for LOTE is difficult for those who do not know target languages.
When we generate synthetic data, we should prepare at least three elements:
1) Word vocabulary (word corpus or lexicon). 
2) Characters that compose words.
3) Font to render words.
However, for people who do not know LOTE, preparing three elements for LOTE is difficult;
difficult to decide appropriate words, characters, and fonts.
In addition, some languages have specific rules to generate their languages.
For example, Arabic texts have two different features from English:
1) They are written from right to left.
2) Some of the Arabic characters change their shape depending on their surrounding characters.
These factors make generating synthetic data of Arabic difficult for those who do not know Arabic.

In this case, when we have to recognize LOTE but generating synthetic data for LOTE is difficult, we need to achieve competitive performance with a few real labels.

\PAR{Need \#3: Current public synthetic data can be inappropriate to datasets other than STR benchmark datasets}
In addition to evaluation on benchmark datasets in \S\ref{exp:benchmark}, we also evaluate the other seven datasets: COCO, RCTW, Uber, ArT, LSVT, MLT19, and ReCTS.
Table~\ref{sup-tab:eval-seven-addition} shows the results.
Baseline-real has higher accuracy than Baseline-synth (Avg. 58.0\% vs. 52.3\% for CRNN and 69.2\% vs. 60.4\% for TRBA).
Our best setting PR successfully improve Baseline-real (Avg. +5.8\% for CRNN and +3.1\% for TRBA).
These results indicate that \textit{fewer real data (Real-L) can be more appropriate to evaluation sets of these seven datasets, rather than current synthetic data (MJ+ST)}.
We presume that because some of the fewer real data (Real-L) derive from the same domain with evaluation sets of seven datasets, Baseline-real has higher accuracy.
Namely, using only fewer real data collected from the target domain can be more significant than using large synthetic data from the other domain.
This indicates that studies on STR with fewer labels are necessary, fully exploiting few real data derived from the target domain.

\PAR{Detailed survey of the literatures of STR with fewer real labels}
As mentioned in \S\ref{sec:introduction}, after emerging large synthetic data~\cite{MJSynth}(2014), the study of STR with fewer labels has decreased.
Thus, there are only a few studies on STR with fewer labels for five years.
Searching for studies on STR with fewer labels is difficult.
In this work, we struggle to search for studies on STR with fewer labels via the widely-used search engine: Arxiv Sanity\footnote{http://www.arxiv-sanity.com/}.

Specifically, we search studies with query ``text recognition'' from 2016 to 2020.
The number of searched papers is 131 papers.
We manually check the contents of all of them, and find four papers related to STR with fewer labels; they only use real data rather than synthetic data.
Three of them are published three or four years ago, and do not use deep learning based methods; 
Mishra \etal~\cite{mishra2016enhancing}(2016) uses the conditional random field model and support vector machine.
Lou \etal~\cite{NIPS2016_shape}(2016) uses the generative shape model and support vector machine.
Bhunia \etal~\cite{Bhunia_2017}(2017) uses the hidden Markov model and support vector machine.
However, it is difficult to compare them with recent state-of-the-art methods.
Because their evaluation data are different from the recent STR benchmark: they all do not evaluate on IC15, SP, and CT.
In addition, they use lexicons for calculating accuracy.
In general, the accuracy calculated with the lexicon has higher accuracy than the one calculated without the lexicon.
Thus, comparing them is unfair.
Recent state-of-the-art methods present both accuracies calculated with lexicons and calculated without lexicons~\cite{CRNN, ASTER, ESIR, Mask-textspotter, ScRN, textscanner} or do not present the accuracy calculated with lexicons~\cite{TRBA, DAN, SE-ASTER, robustscanner, plugnet}.
Therefore, the former can be compared with the three methods~\cite{mishra2016enhancing, NIPS2016_shape, Bhunia_2017} but the latter cannot.
The process of using lexicon to calculate accuracy is as follows:
1) IIIT, SVT, and SP have predefined lexicons which contain a target word and some candidate words.
2) Transform the output word from the model into the closet word in predefined lexicons.
3) Calculate accuracy between transformed one and the target word.

Another paper uses deep learning based methods~\cite{ArT-won-onlyReal}(2019).
The paper addresses recognizing English and Chinese texts.
However, strictly speaking, the paper describes scene text spotting (STS), which is the combination of scene text detection (STD) and STR, rather than sole STR.
Because they do not describe the result of STR in their paper, it is difficult to compare their method with other STR methods.
From the results of STS, we can presume that their detector has great performance but their recognizer has a less competitive performance to other English STR methods.
Because they won 1st for STS in the ArT competition (English only), but they recorded 15th of 23 teams for STR in the ArT competition (English only)\footnote{https://rrc.cvc.uab.es/?ch=14\&com=evaluation\&task=2}.
However, this is not a concrete comparison but only our presumption.
We need the results on STR benchmark datasets rather than the results of STS to compare their method with other STR methods.

It is currently difficult to search for studies on STR with fewer labels and compare those studies with state-of-the-art methods.
We hope that our study becomes the appropriate study on STR with fewer labels that can be easily searched and compared.

\clearpage

\section{STR Datasets - Details and More Examples}\label{sup:data}
In this section, we describe the details of our preprocessing process in \S\ref{sec:consoli}.
Furthermore, we illustrate more examples of real data.

\subsection{Preprocessing Real Datasets}\label{sup:preprocessing}

\PAR{Consolidating real datasets}
We collect SVT and IIIT from their web page\footnote{http://vision.ucsd.edu/~kai/svt/}\textsuperscript{,}\footnote{https://cvit.iiit.ac.in/research/projects/cvit-projects/the-iiit-5k-word-dataset}.
However, the labels of training set of SVT and IIIT are case-insensitive; they are all uppercase alphabet.
We use case sensitive data of SVT and IIIT corrected by \cite{long2020unrealtext}\footnote{https://github.com/Jyouhou/Case-Sensitive-Scene-Text-Recognition-Datasets}.
We collect IC13, IC15, COCO, ArT, LSVT, MLT19, and ReCTS from the web page of ICDAR competition\footnote{https://rrc.cvc.uab.es/}.
We download RCTW and Uber from their web page\footnote{https://rctw.vlrlab.net/dataset}\textsuperscript{,}\footnote{https://s3-us-west-2.amazonaws.com/uber-common-public/ubertext/index.html}.

\begin{figure}
\centering
\includegraphics[width=0.48\linewidth, height=2.5cm]{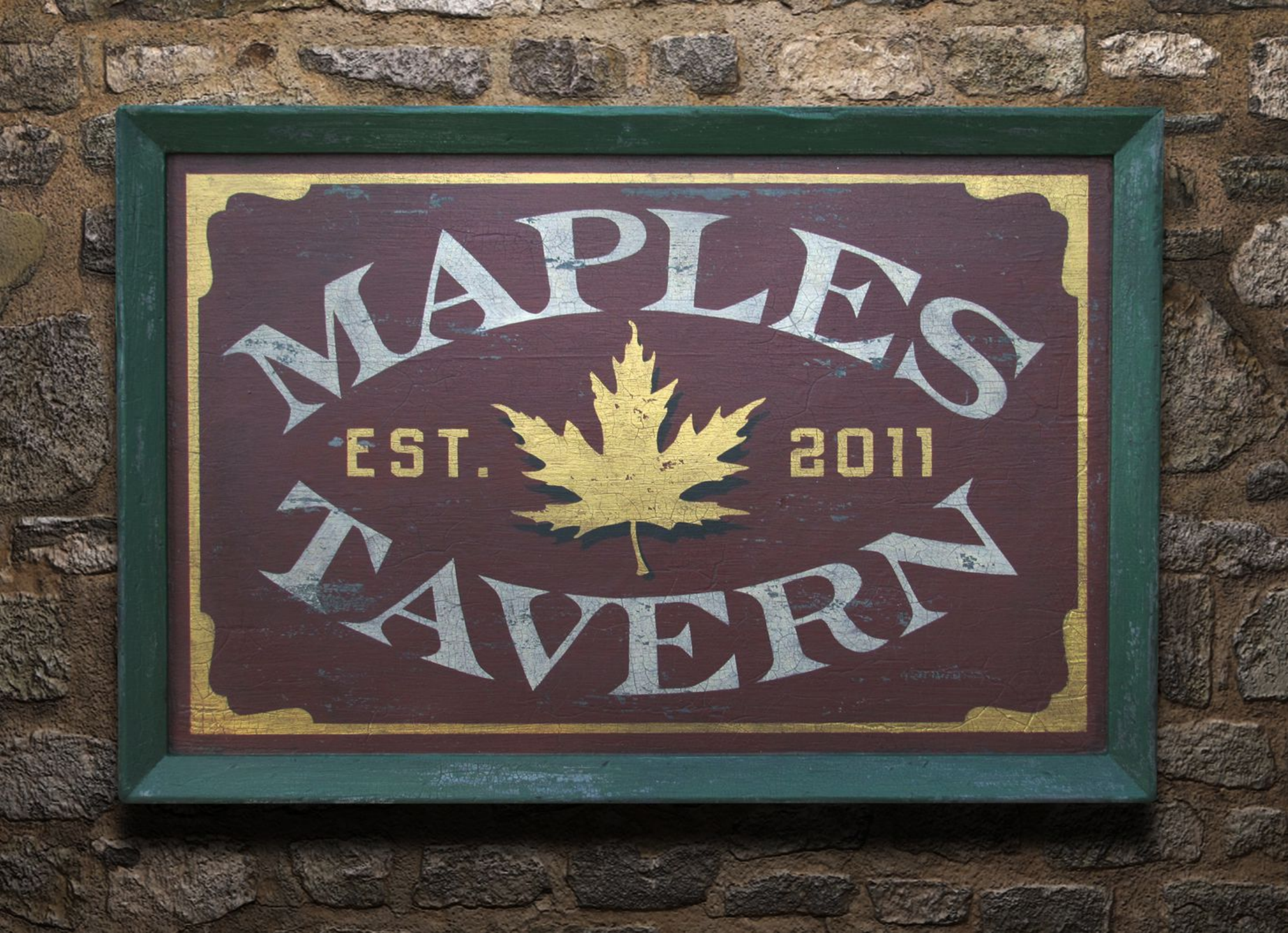}
\hspace{0.1mm}
\includegraphics[width=0.48\linewidth, height=2.5cm]{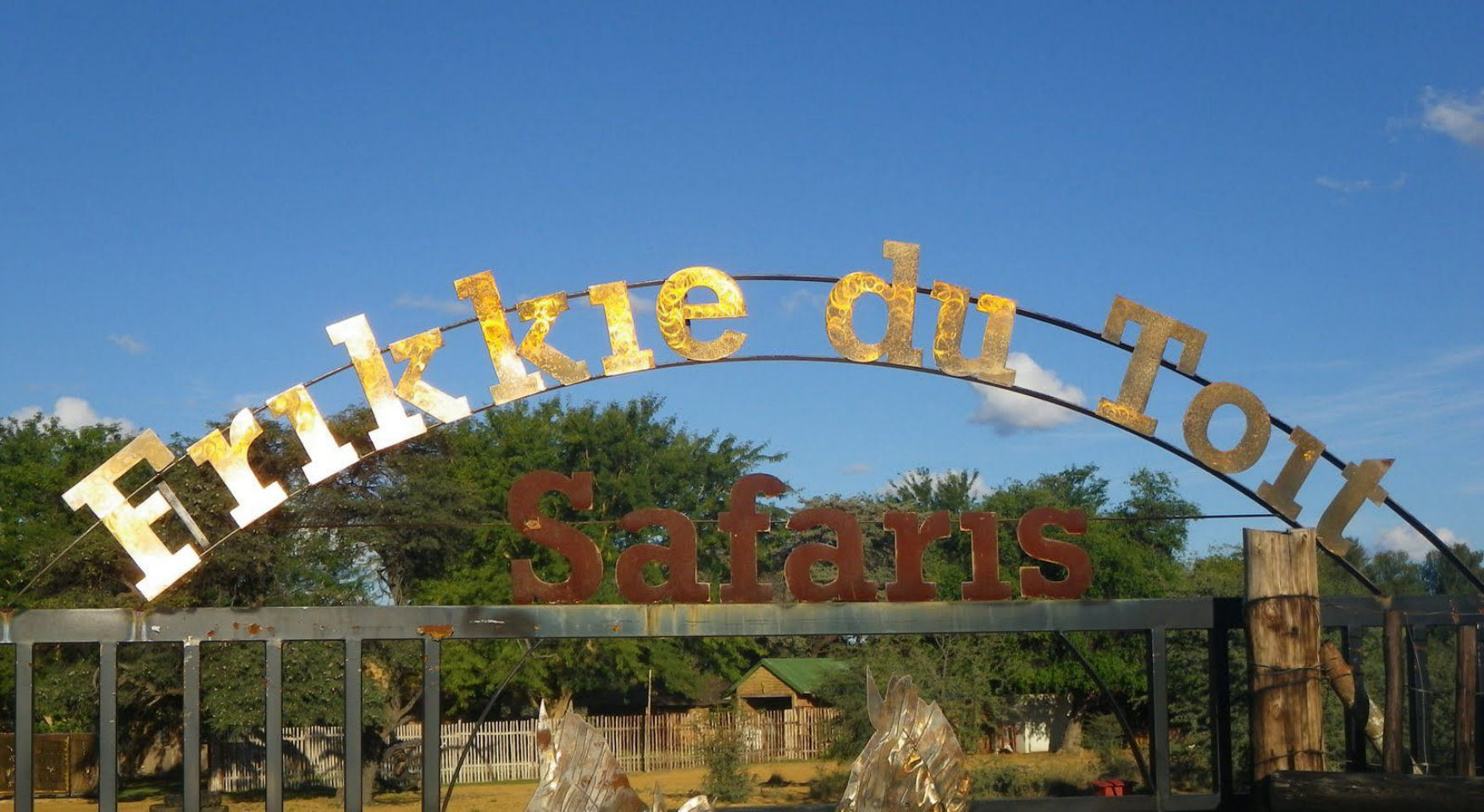}
  \vspace{-1mm}
  \caption{
    Duplicated scene images. 
    These images are found in both the training set of ArT and the evaluation set of CT.
  }
  \label{sup-fig:ArT-CT}
\end{figure}

\PAR{Excluding duplication between datasets}
If we do not pay attention the duplication between datasets, we might use the training set that includes some of the evaluation set.
To avoid this duplication, we investigate the duplication between all training set and all evaluation set.
Specifically, we investigate whether the scene images of training sets match the scene images of evaluation sets.
As a result, we find that the training set of Art and the evaluation set of CT have 27 duplicated scene images and 122 duplicated word boxes, as shown in Figure~\ref{sup-fig:ArT-CT}.
We exclude them for a fair comparison.
The journal version paper of Total-Text~\cite{totaltext} indicates that some of the word boxes in Total-Text are duplicated in CT. 
ArT includes Total-Text, and therefore some of the word boxes in ArT can be also duplicated in CT.
However, we do not know how many Total-Text images are in the training set of ArT.
Thus, we investigate the duplicated word boxes by comparing scene images of ArT and CT.
For another example, according to \cite{TRBA}, 215 word boxes are duplicated in the training set of IC03 and the evaluation set of IC13.
This is one of the reasons why we exclude IC03; the other reason is that IC13 inherits most of IC03 data.

\begin{figure}
\centering
    \begin{subfigure}{0.485\linewidth} \centering
     \includegraphics[width=0.97\linewidth, height=2.5cm]{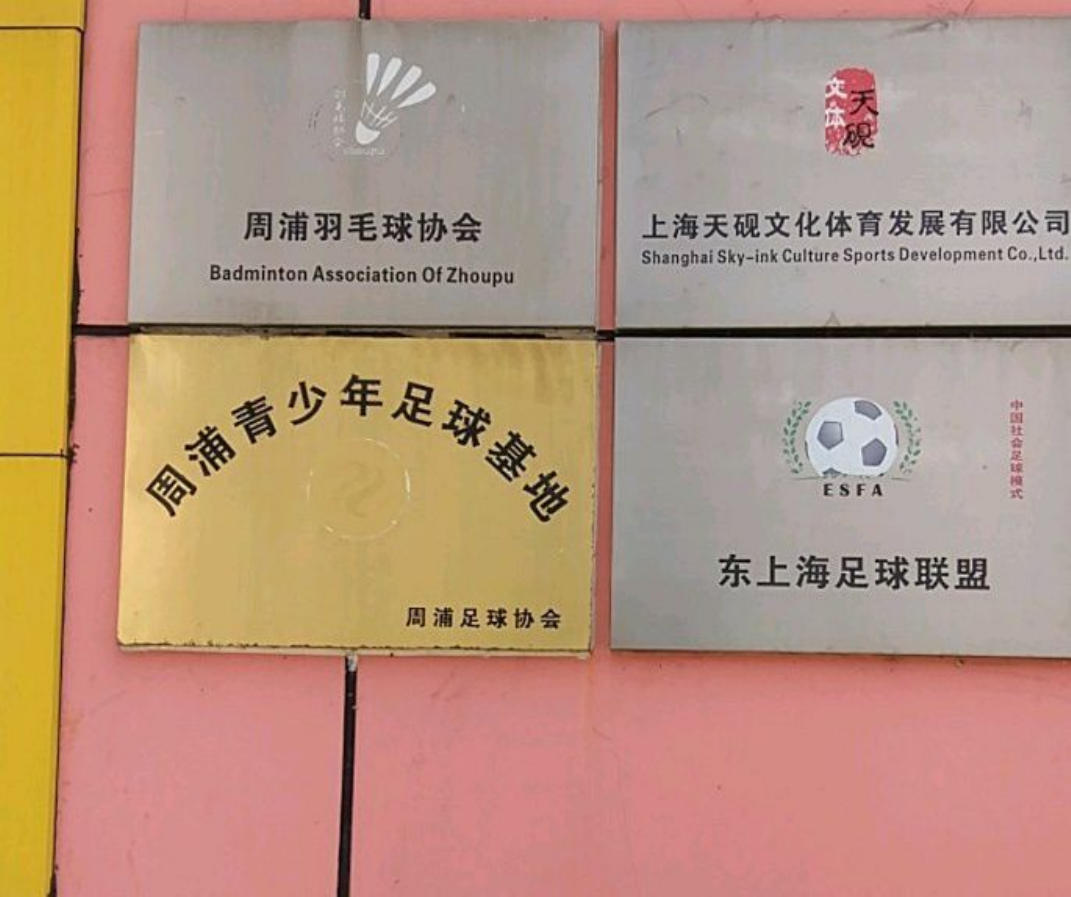}
     \vspace{-1mm}
     \caption{Image from ArT}
    \end{subfigure}
    \begin{subfigure}{0.485\linewidth} \centering
     \includegraphics[width=0.97\linewidth, height=2.5cm]{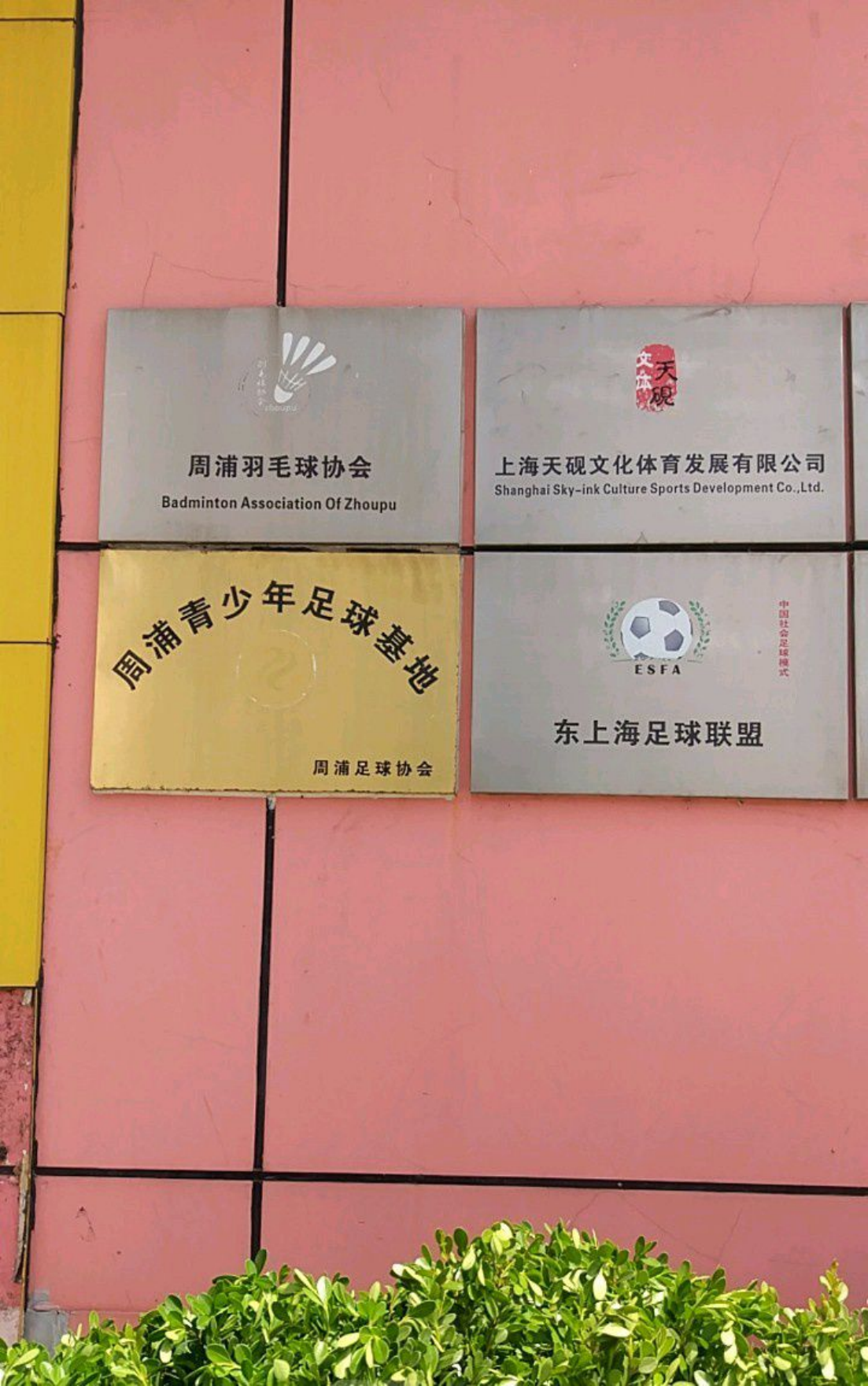}
     \vspace{-1mm}
     \caption{Image from LSVT}
    \end{subfigure}
  \vspace{-2mm}
  \caption{
    Similar scene images. 
    These images are found in both training sets of ArT and LSVT.
    They are different images but contain the same texts; the text regions of ArT are usually more enlarged than LSVT.
  }
  \label{sup-fig:ArT-LSVT}
\end{figure}

According to \cite{ArT}, ArT includes the subset of LSVT.
We cannot find duplicated images by matching scene images because they are slightly different, as shown in Figure~\ref{sup-fig:ArT-LSVT}.
In this case, we investigate duplication by matching labels of scene images.
As a result, we find that the training set of Art and the training set of LSVT have 814 similar scene images (433 scene images for English) and 4,578 similar word boxes (861 word boxes for English).

However, since labels of scene images for the evaluation set of ArT and LSVT are not provided, we cannot find duplication by matching labels of scene images.
Thus, it is difficult to exclude the duplication between the evaluation set of ArT and the training set of LSVT.
If there are duplicated images between them, comparing the accuracy of ArT between a method that uses the training set of LSVT and another method that does not use the training set of LSVT can be unfair.

In the case of evaluation on LSVT, in our experiments, we split the training set of LSVT into training, validation, and evaluation set.
We exclude the duplicated word boxes between the training set of ArT and the evaluation set (after splitting) of LSVT; exclude 74 word boxes in the evaluation set of LSVT.
In our experiments, we would like to use more real data for training and validation; thus, we exclude the duplicated word boxes in the evaluation set of LSVT rather than exclude training and validation sets of ArT.

\PAR{Collecting only English words} 
In this study, we take only English and symbols.
Specifically, we exclude Chinese characters in RCTW, ArT, LSVT, and ReCTS.
For MLT, all word labels have ``a script label'' representing the language of each word label.
We exclude the words whose script label is Arabic, Chinese, Japanese, Korean, Bangla, or Hindi. 

\PAR{Excluding don't care symbol}
Don't care symbol ``*'' or ``\#'' is usually included in recent public real data: RCTW, Uber, ArT, LSVT, MLT19, and ReCTS.
We do not conduct this filtering on SVT, IIIT, IC13, IC15, and COCO.

We do not exclude the texts contains ``\#'' symbol. 
Instead, we only exclude the texts is ``\#'', ``\#\#'', ``\#\#\#'', or ``\#\#\#\#''.
Therefore, we can train ``\#'' symbol with other characters.
In the case of MLT19, we only exclude the texts is ``\#\#\#'' or ``\#\#\#\#''.

In contrast to ``\#'' symbol, we exclude the image whose label contains ``*'' symbol in Uber.
About half of Uber images (149K of 285K training images) contain ``*'' symbol. 
Although some of them are sufficiently readable, they contain ``*'' symbol.
Thus, we exclude them because they can be the noise of data.

\PAR{Excluding vertical or $\pm$ 90 degree rotated texts}
As described in \S\ref{sec:preprocessing}, we mainly focus on horizontal texts and thus exclude vertical texts.
For the characters such as ``1, i, j, l, t'' and the words such as ``it'', their height usually are greater than their width. 
To avoid excluding these characters and words, we only exclude images \textit{whose texts have more than two characters} and whose height is greater than the width.

\PAR{Splitting each real dataset into training, validation, and evaluation sets}
Table~\ref{sup-tab:split} shows whether each dataset originally has training, validation, and evaluation sets.
Some datasets do not have them, and thus we split the training set of the dataset.
For example, RCTW, LSVT, and MLT19 only have a training set. 
Thus we split the training set of each dataset into training, validation, and evaluation sets with ratios of 80\%, 10\%, and 10\%, respectively.
The evaluation sets of RCTW, LSVT, and MLT19 have been released, but the evaluation sets only contain images and do not contain labels.
Therefore, we cannot evaluate their evaluation sets, and thus RCTW, LSVT, and MLT19 are considered not to have the evaluation set.

In the other case, SVT, IIIT, IC13, IC15, ArT, and ReCTS do not have a validation set, and thus we split the training set of each dataset into training and validation sets with ratios of 90\% and 10\%, respectively.

COCO and Uber originally have training, validation, and evaluation sets. 
Thus we use their original training, validation, and evaluation sets. 
We do not split their training sets.

Table~\ref{sup-tab:split} shows the number of word boxes used in our experiments.
The number is calculated as follows:
1) Conduct preprocessing on the training set of each dataset as described in \S\ref{sec:preprocessing} and \S\ref{sup:preprocessing}.
2) Following the base code~\cite{TRBA}, exclude the images whose texts have more than 25 characters in our experiments.

\begin{table}[t] 
    \begin{center}
        \begin{adjustbox}{width=0.975\linewidth}
        \begin{tabular}{@{}llcccrrr@{}}
            \toprule
            & & \multicolumn{3}{c}{Check for inclusion} & \multicolumn{3}{c}{\# of word boxes} \\ \cmidrule(l){3-5}\cmidrule(l){6-8}
            \multicolumn{2}{l}{Dataset} & Train. & Valid. & Eval. & Train. & Valid. & Eval. \\
            \midrule
            \multicolumn{4}{l}{\textbf{Real labeled datasets (Real-L)}} \\ 
            ~~ & SVT  & $\checkmark$ & $-$ & $\checkmark$ & 231 & 25 & 647 \\
            ~~ & IIIT & $\checkmark$ & $-$ & $\checkmark$ & 1,794 & 199 & 3,000 \\
            ~~ & IC13 & $\checkmark$ & $-$ & $\checkmark$ & 763 & 84 & 1,015 \\
            ~~ & IC15 & $\checkmark$ & $-$ & $\checkmark$ & 3,710 & 412 & 2,077 \\ 
            ~~ & COCO & $\checkmark$ & $\checkmark$ & $\checkmark$ & 39K & 9,092 & 9,823 \\ 
            ~~ & RCTW & $\checkmark$ & $-$ & $-$ & 8,186 & 1,026 & 1,029 \\
            ~~ & Uber & $\checkmark$ & $\checkmark$ & $\checkmark$ & 92K & 36K & 80K \\
            ~~ & ArT & $\checkmark$ & $-$ & $\checkmark$ & 29K & 3,202 & 35K \\
            ~~ & LSVT & $\checkmark$ & $-$ & $-$ & 34K & 4,184 & 4,133 \\
            ~~ & MLT19 & $\checkmark$ & $-$ & $-$ & 46K & 5,689 & 5,686\\
            ~~ & ReCTS & $\checkmark$ & $-$ & $\checkmark$ & 23K & 2,531 & 2,592 \\
            \arrayrulecolor{lightgray}\hline\arrayrulecolor{black}
            ~~ & Total & $-$ & $-$ & $-$ & 276K & 63K & 146K \\
            \bottomrule
        \end{tabular}
        \end{adjustbox}
    \vspace{-2mm}
    \caption{Number of word boxes used in our experiments, after splitting each real dataset into training, validation, and evaluation sets.}
    \label{sup-tab:split}
    \end{center}
    \vspace{-2mm}
\end{table}

\PAR{Detector for cropping texts in unlabeled scene images}
Unlabeled datasets described in \S\ref{sec:real_unlabel} contain scene images.
The scene images do not have labels indicating text region.
Therefore, we use pretrained scene text detection (STD) model for cropping texts in the scene images.
In this paper, we assume the case when we have to train STR models without synthetic data.
In this case, we cannot train STD model with synthetic data, and thus we use STD model trained only on real data, called BDN~\cite{BDN-ReCTS}.
BDN has two versions; 
1) BDN published in IJCAI~\cite{BDN} use synthetic data (ST~\cite{SynthText}) for training.
2) BDN used for ReCTS competition~\cite{BDN-ReCTS} do not use synthetic data.
We use the second one for cropping texts in unlabeled scene images.

\subsection{More Examples of Public Real Data}
\begin{figure*}[t]
\centering
    \begin{subfigure}{0.33\linewidth} \centering
     \includegraphics[width=0.97\linewidth, height=2.5cm]{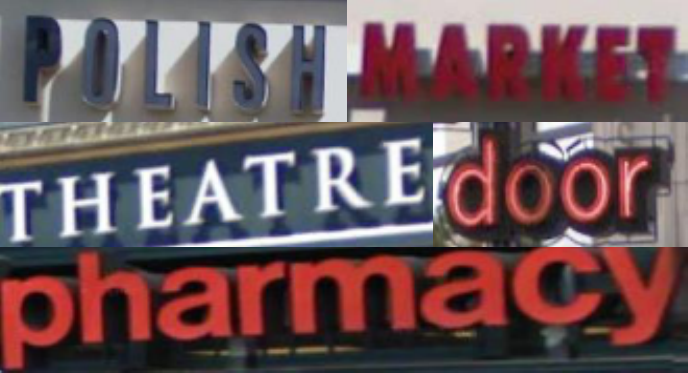}
     \vspace{-1mm}
     \caption{SVT}
    \end{subfigure}
    \begin{subfigure}{0.33\linewidth} \centering
     \includegraphics[width=0.97\linewidth, height=2.5cm]{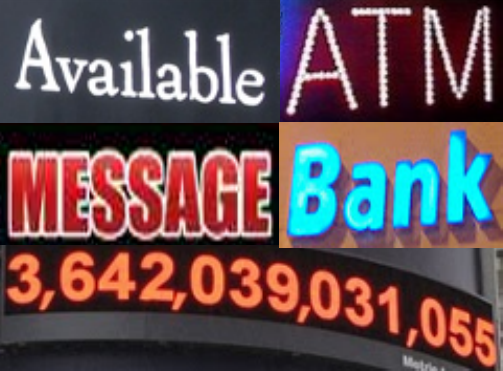}
     \vspace{-1mm}
     \caption{IIIT}
    \end{subfigure}
    \begin{subfigure}{0.33\linewidth} \centering
     \includegraphics[width=0.97\linewidth, height=2.5cm]{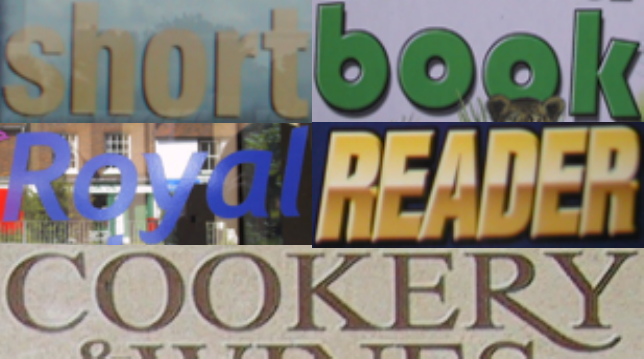}
     \vspace{-1mm}
     \caption{IC13}
    \end{subfigure}
    
    \vspace{1mm}
    \begin{subfigure}{0.33\linewidth} \centering
     \includegraphics[width=0.97\linewidth, height=2.5cm]{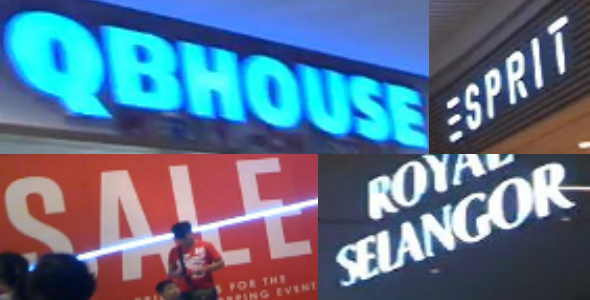}
     \vspace{-1mm}
     \caption{IC15}
    \end{subfigure}
    \begin{subfigure}{0.33\linewidth} \centering
     \includegraphics[width=0.97\linewidth, height=2.5cm]{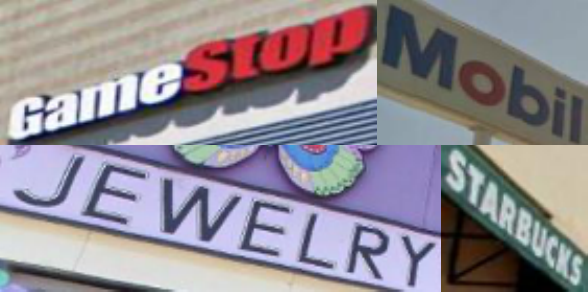}
     \vspace{-1mm}
     \caption{SP}
    \end{subfigure}
    \begin{subfigure}{0.33\linewidth} \centering
     \includegraphics[width=0.97\linewidth, height=2.5cm]{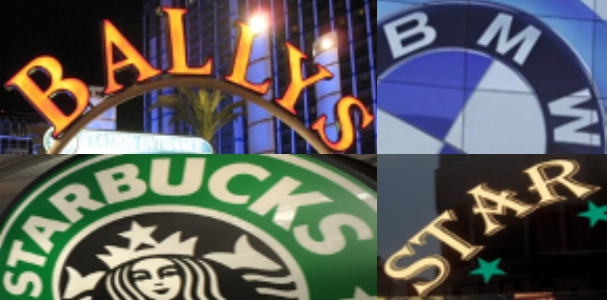}
     \vspace{-1mm}
     \caption{CT}
    \end{subfigure}
\vspace{-2mm}
\caption{Examples of benchmark datasets for evaluation. 
SVT, IIIT, and IC13 are regarded as regular datasets. 
They contain many horizontal texts.
IC15, SP, and CT are regarded as irregular dataets.
They contain many perspective or curved texts.}
\label{sup-fig:benchmark-data}
\end{figure*}

\begin{figure*}[t]
\centering
    \begin{subfigure}{0.485\linewidth} \centering
     \includegraphics[width=0.485\linewidth, height=2.5cm]{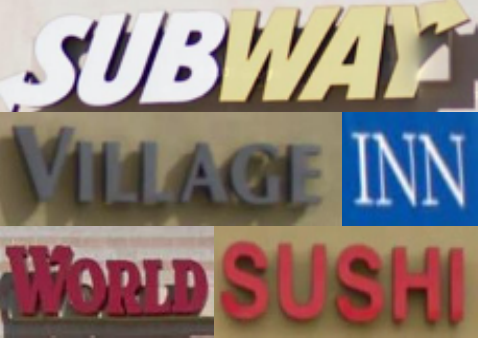}
     \includegraphics[width=0.485\linewidth, height=2.5cm]{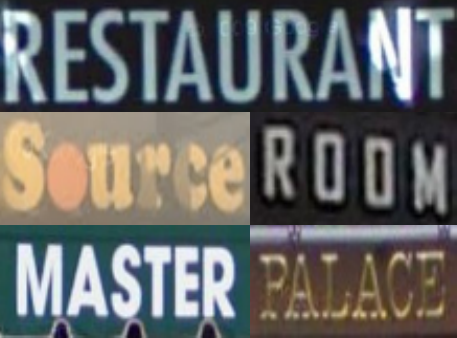}
     \vspace{-1mm}
     \caption{Year 2011: SVT}
    \end{subfigure}
    \begin{subfigure}{0.485\linewidth} \centering
     \includegraphics[width=0.485\linewidth, height=2.5cm]{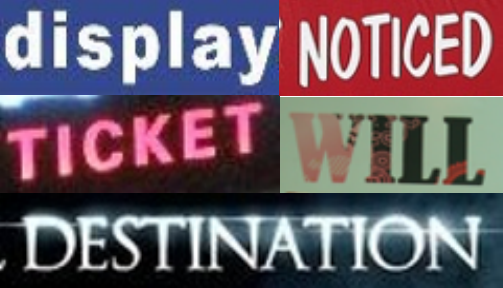}
     \includegraphics[width=0.485\linewidth, height=2.5cm]{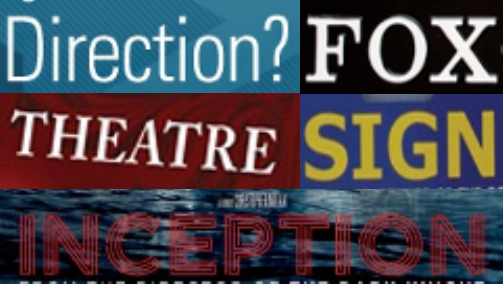}
     \vspace{-1mm}
     \caption{Year 2013: IIIT}
    \end{subfigure}
    
    \vspace{1mm}
    \begin{subfigure}{0.485\linewidth} \centering
     \includegraphics[width=0.485\linewidth, height=2.5cm]{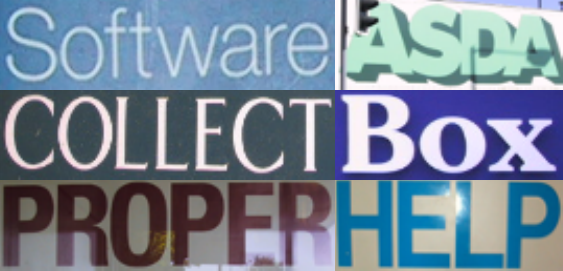}
     \includegraphics[width=0.485\linewidth, height=2.5cm]{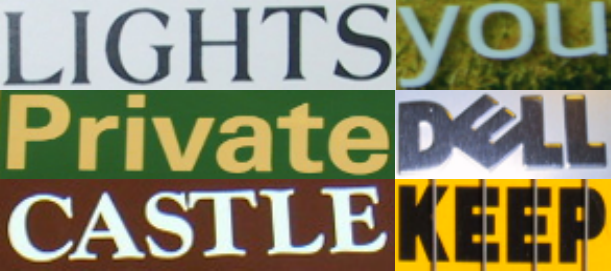}
     \vspace{-1mm}
     \caption{Year 2013: IC13}
    \end{subfigure}
    \begin{subfigure}{0.485\linewidth} \centering
     \includegraphics[width=0.485\linewidth, height=2.5cm]{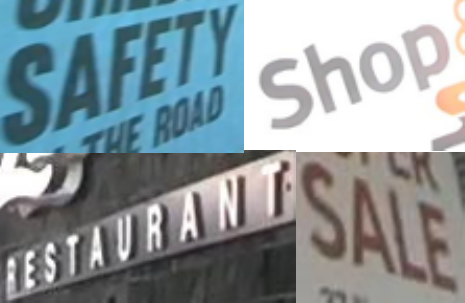}
     \includegraphics[width=0.485\linewidth, height=2.5cm]{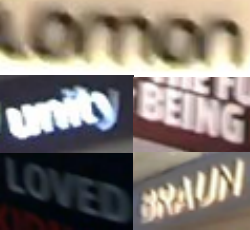}
     \vspace{-1mm}
     \caption{Year 2015: IC15}
    \end{subfigure}
\vspace{-2mm}
\caption{Examples of public real datasets for Year 2011 (SVT), Year 2013 (IIIT and IC13), and Year 2015 (IC15).
SVT, IIIT, and IC13 contains many horizontal texts.
IC15 contains many perspective or blurry texts.}
\label{sup-fig:Y2011-2015}
\end{figure*}

\begin{figure*}[t]
\centering
    \begin{subfigure}{\linewidth} \centering
     \includegraphics[width=0.235\linewidth, height=2.5cm]{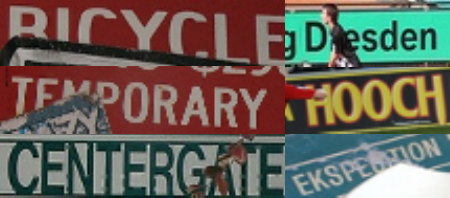}
     \includegraphics[width=0.235\linewidth, height=2.5cm]{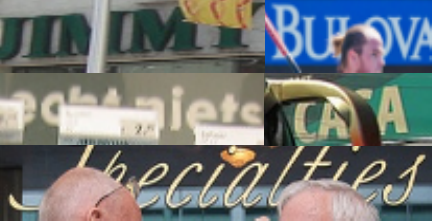}
     \hspace{0.7mm}
     \includegraphics[width=0.235\linewidth, height=2.5cm]{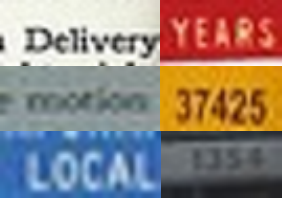}
     \includegraphics[width=0.235\linewidth, height=2.5cm]{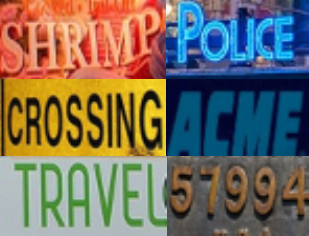}
     \vspace{-1mm}
     \caption{COCO contains many occluded or low-resolution texts.}
    \end{subfigure}
    
    \vspace{1mm}
    \begin{subfigure}{0.485\linewidth} \centering
     \includegraphics[width=0.485\linewidth, height=2.5cm]{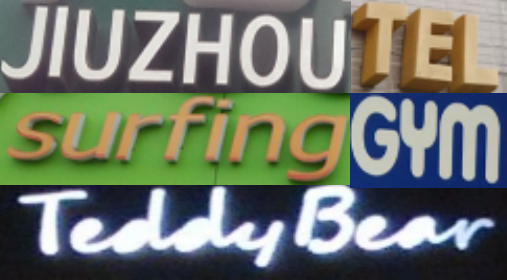}
     \includegraphics[width=0.485\linewidth, height=2.5cm]{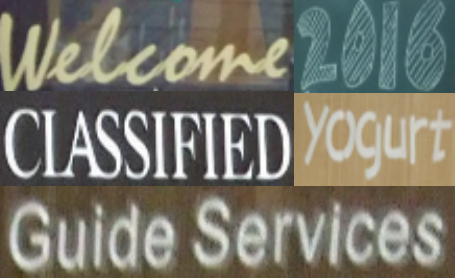}
     \vspace{-1mm}
     \caption{RCTW: Most images are collected in China.}
    \end{subfigure}
    \begin{subfigure}{0.485\linewidth} \centering
     \includegraphics[width=0.485\linewidth, height=2.5cm]{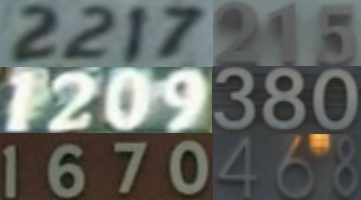}
     \includegraphics[width=0.485\linewidth, height=2.5cm]{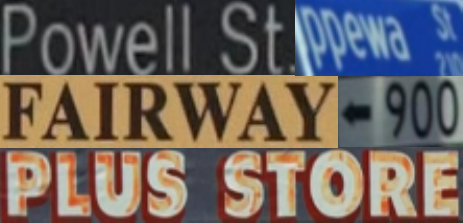}
     \vspace{-1mm}
     \caption{Uber contains many house number or signboard texts.}
    \end{subfigure}
\vspace{-2mm}
\caption{Examples of public real datasets for Year 2017: COCO, RCTW, and Uber.}
\label{sup-fig:Y2017}
\end{figure*}

\begin{figure*}[t]
\centering
    \begin{subfigure}{0.97\linewidth} \centering
     \includegraphics[width=0.32\linewidth, height=2.5cm]{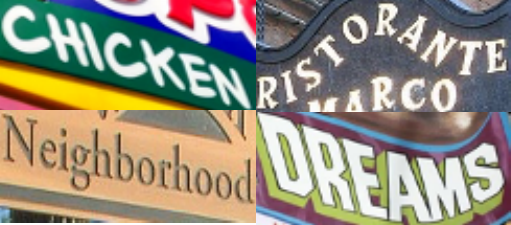}
     \hspace{1mm}
     \includegraphics[width=0.32\linewidth, height=2.5cm]{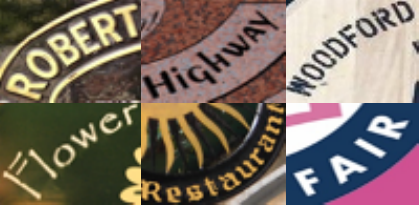}
     \hspace{1mm}
     \includegraphics[width=0.32\linewidth, height=2.5cm]{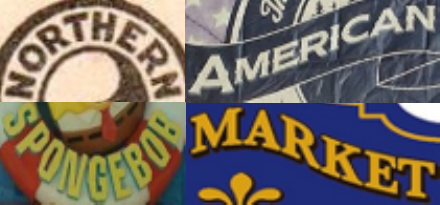}
     \vspace{-1mm}
     \caption{ArT contains many arbitrary-shaped texts: rotated, perspective or curved texts.}
    \end{subfigure}
    
    \vspace{1mm}
    \begin{subfigure}{0.97\linewidth} \centering
     \includegraphics[width=0.32\linewidth, height=2.5cm]{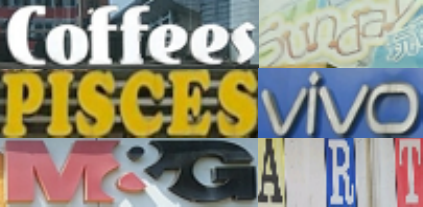}
     \hspace{1mm}
     \includegraphics[width=0.32\linewidth, height=2.5cm]{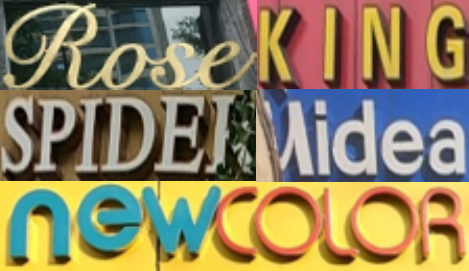}
     \hspace{1mm}
     \includegraphics[width=0.32\linewidth, height=2.5cm]{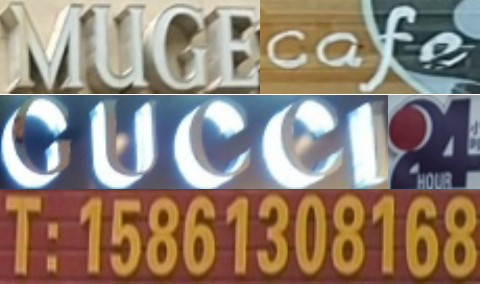}
     \vspace{-1mm}
     \caption{LSVT is collected from streets in China.}
    \end{subfigure}
    
    \vspace{1mm}
    \begin{subfigure}{0.97\linewidth} \centering
     \includegraphics[width=0.32\linewidth, height=2.5cm]{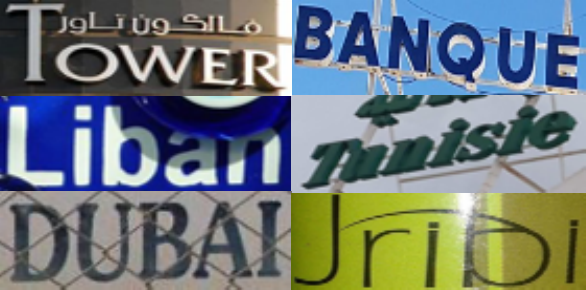}
     \hspace{1mm}
     \includegraphics[width=0.32\linewidth, height=2.5cm]{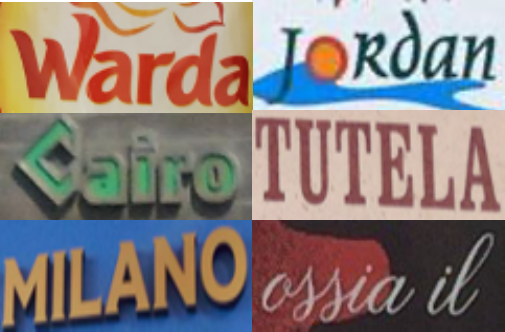}
     \hspace{1mm}
     \includegraphics[width=0.32\linewidth, height=2.5cm]{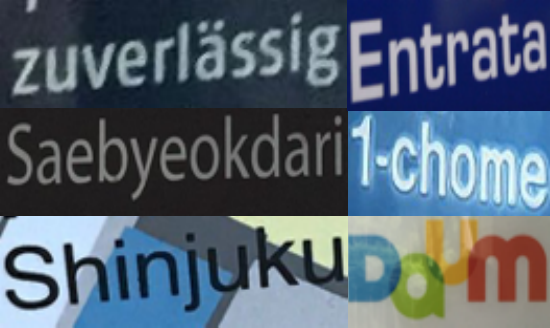}
     \vspace{-1mm}
     \caption{MLT19 is collected to recognize multi-lingual texts.}
    \end{subfigure}
    
    \vspace{1mm}
    \begin{subfigure}{0.97\linewidth} \centering
     \includegraphics[width=0.32\linewidth, height=2.5cm]{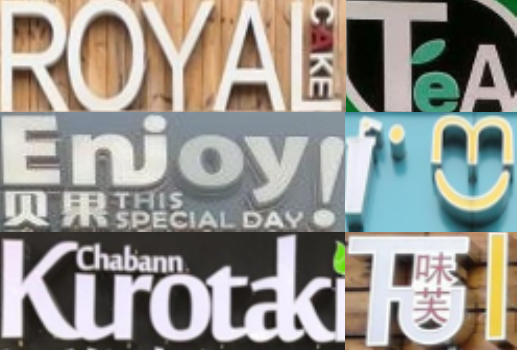}
     \hspace{1mm}
     \includegraphics[width=0.32\linewidth, height=2.5cm]{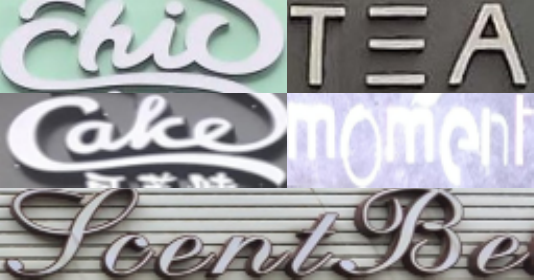}
     \hspace{1mm}
     \includegraphics[width=0.32\linewidth, height=2.5cm]{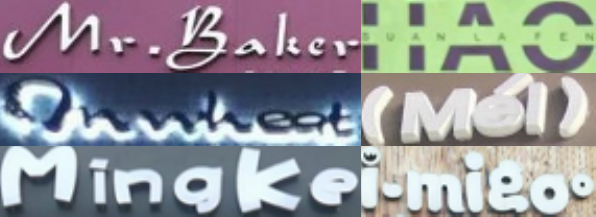}
     \vspace{-1mm}
     \caption{ReCTS contains texts arranged in various layouts or texts written in difficult fonts.}
    \end{subfigure}
\vspace{-2mm}
\caption{Examples of public real datasets for Year 2019: ArT, LSVT, MLT19, and ReCTS.}
\label{sup-fig:Y2019}
\end{figure*}

\begin{figure*}[t]
\centering
    \begin{subfigure}{0.33\linewidth} \centering
     \includegraphics[width=0.97\linewidth, height=3cm]{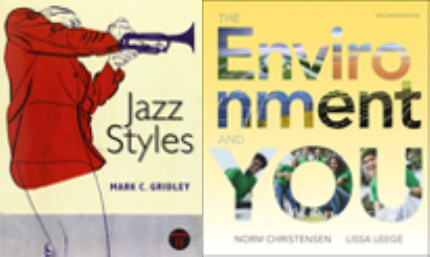}
     \vspace{-1mm}
     \caption{Book32: scene images.}
    \end{subfigure}
    \begin{subfigure}{0.33\linewidth} \centering
     \includegraphics[width=0.97\linewidth, height=3cm]{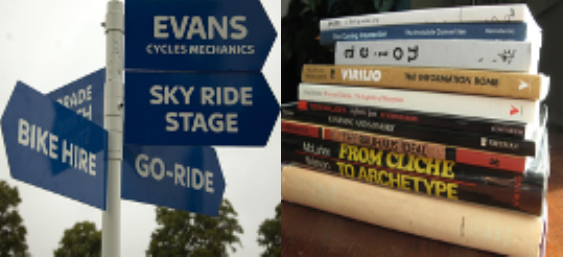}
     \vspace{-1mm}
     \caption{TextVQA: scene images.}
    \end{subfigure}
    \begin{subfigure}{0.33\linewidth} \centering
     \includegraphics[width=0.97\linewidth, height=3cm]{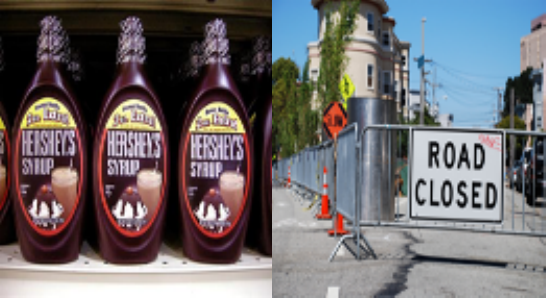}
     \vspace{-1mm}
     \caption{ST-VQA: scene images.}
    \end{subfigure}
    
    \vspace{1mm}
    \begin{subfigure}{0.33\linewidth} \centering
     \includegraphics[width=0.97\linewidth, height=2.5cm]{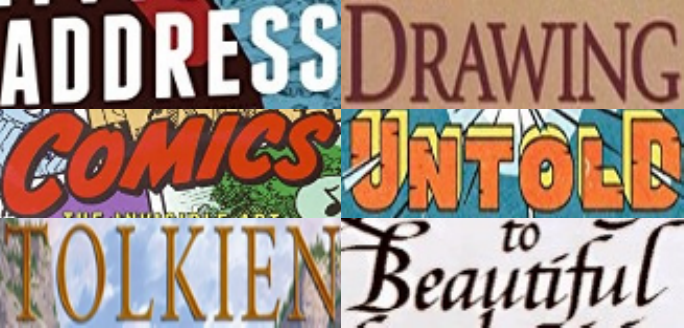}
     \vspace{-1mm}
     \caption{Book32: word boxes after cropping.}
    \end{subfigure}
    \begin{subfigure}{0.33\linewidth} \centering
     \includegraphics[width=0.97\linewidth, height=2.5cm]{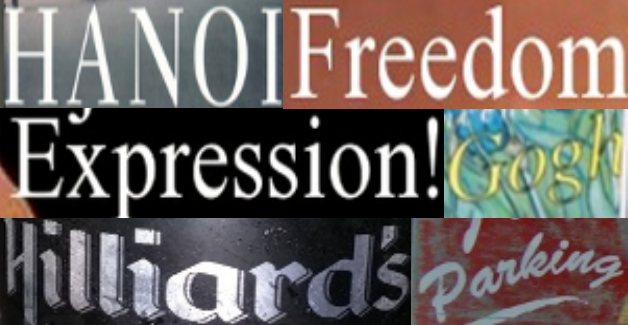}
     \vspace{-1mm}
     \caption{TextVQA: word boxes after cropping.}
    \end{subfigure}
    \begin{subfigure}{0.33\linewidth} \centering
     \includegraphics[width=0.97\linewidth, height=2.5cm]{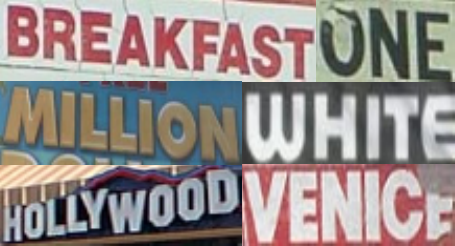}
     \vspace{-1mm}
     \caption{ST-VQA: word boxes after cropping.}
    \end{subfigure}
\vspace{-2mm}
\caption{Examples of unlabeled datasets: Book32, TextVQA, and ST-VQA. 
We show their scene images and word boxes after cropping.}
\label{sup-fig:unlabel}
\end{figure*}

Recently, various types of texts have been accumulated.
They can improve the robustness of STR models.
We present more examples of public real data.
Figure~\ref{sup-fig:benchmark-data} shows the examples of benchmark datasets for evaluation, as described in \S\ref{sec:benchmark_data}.
Figure~\ref{sup-fig:Y2011-2015}, \ref{sup-fig:Y2017}, and \ref{sup-fig:Y2019} are the extended versions of Figure~\ref{fig:dataset} in \S\ref{sec:consoli}.
Figure~\ref{sup-fig:Y2011-2015} shows examples of accumulated real labeled data for Year 2011, 2013, and 2015.
Figure~\ref{sup-fig:Y2017}, and \ref{sup-fig:Y2019} show examples of that of Year 2017 and 2019, respectively.
Figure~\ref{sup-fig:unlabel} shows examples of unlabeled scene images and word boxes after cropping.

\clearpage

\clearpage

\section{STR With Fewer Labels - Details}\label{sup:method}
We describe more details of STR models and semi- and self-supervised learning in \S\ref{sec:method}.

\subsection{Description of STR Models}
We select two models in order to represent that our experimental results are not limited to a specific STR model.
We adopt two widely-used STR models in STR benchmark repository~\cite{TRBA}\footnote{https://github.com/clovaai/deep-text-recognition-benchmark}: CRNN~\cite{CRNN} and TRBA~\cite{TRBA}.

CRNN, the abbreviation of Convolutional-Recurrent Neural Network, is the first model that combines convolutional neural networks (CNNs) and recurrent neural networks (RNNs) for STR.
CRNN is regarded as a basic and lightweight STR model.
Thus, CRNN is usually selected as one of the baseline models.
Also, CRNN is sometimes adopted to the STR part of scene text spotting that is the combination of scene text detection and recognition~\cite{TextboxAAAI17-CRNNusedforSTS,Textbox++2018TIP-CRNNusedforSTS,liu2020abcnet-CRNNusedforSTS}.
CRNN consists of 7 layers of VGG-like CNNs, 2 layers of BiLSTM, and CTC decoder.

TRBA, the abbreviation of TPS-ResNet-BiLSTM-Attention, has the best performance in STR benchmark repository~\cite{TRBA}.
TRBA is also usually selected as one of the baseline models~\cite{scatter,Xu_2020_CVPR-WhatMachines-TRBAused, PGT-RealSynthMix3, patel2020learning-ECCV-TRBAused, xu2020_ACMMM-adversarial-TRBAused}.
TRBA consists of 4 layers of CNNs for TPS, 29 layers of ResNet, 2 layers of BiLSTM, and an attention decoder.
According to \cite{TRBA}, TRBA is created by composing existing modules used in prior works.
Thus, TRBA can be regarded as a variant of RARE~\cite{RARE}.
RARE is a well-known architecture that introduces an image transformation module into STR models.
The main differences between TRBA and RARE are as follows:
TRBA uses 
1) ResNet instead of VGG-like CNNs.
2) BiLSTM instead of GRU.
TRBA can be also regarded as a variant of another widely-used baseline model called ASTER~\cite{ASTER} because ASTER is the successor of RARE.
The main differences between TRBA and ASTER are as follows:
1) Details of TPS.
2) ASTER uses a bidirectional attention decoder.

\subsection{Objective Function}
We use labeled training set $\mathcal{D}_l = \{(x_1,y_1), ..., (x_n,y_n)\}$ and unlabeled training set $\mathcal{D}_u  = \{u_1, ..., u_m\}$ to calculate object function, where $x$ is the labeled training image, $y$ is the word label, and $u$ is the unlabeled training image.

\PAR{Objective function for STR}
STR models are trained by minimizing the objective function as follows,
\begin{equation}
\mathcal{L}_{\mathrm{recog}}=-\frac{1}{|\mathcal{D}_l|}\sum_{(x,y) \in \mathcal{D}_l}\log p(y|x)
\end{equation}
where $p(y|x)$ is the conditional probability of word label.

\PAR{Objective function for semi-supervised learning}
Pseudo-Label (PL)~\cite{lee2013pseudo} uses pseudolabeled training set $\mathcal{D}_s = \{(u_1,s_1), ..., (u_m,s_m)\}$ for training, where $s$ is the pseudolabel of $u$.
The object function of PL calculated as follows,
\begin{equation}
\mathcal{L} = \mathcal{L}_{\mathrm{recog}} -\frac{1}{|\mathcal{D}_s|} \sum_{(u,s) \in \mathcal{D}_s}\log p(s|u)
\end{equation}
where $p(s|u)$ is the conditional probability of the pseudolabel.

Mean Teacher (MT)~\cite{tarvainen2017mean} uses the object function as follows,
\begin{equation}
\mathcal{L} = \mathcal{L}_{\mathrm{recog}} + \alpha \frac{1}{|\mathcal{D}_l|+|\mathcal{D}_u|}  \sum_{[x;u]}\mathrm{MSE}(f_{\theta}([x;u]^\eta), f_{\theta'}([x;u]^{\eta'}))
\end{equation}
where $[x;u]$ denotes the concatenation of $x$ and $u$. 
$\eta$ and $\eta'$ denote two random augmentations. 
$[x;u]^\eta$ and $[x;u]^{\eta'}$ are the images augmented by $\eta$ and $\eta'$, respectively. 
$\mathrm{MSE}$ denotes the mean squared error, $f_{\theta}$ is the student model, $f_{\theta'}$ is the teacher model, and $\alpha$ is coefficient of MSE loss.

\PAR{Objective function for self-supervised learning}
The pretext task of RotNet~\cite{RotNet} is conducted by minimizing the objective function as follows,
\begin{equation}
\mathcal{L}_{\mathrm{Rot}}=-\frac{1}{|R|}\frac{1}{|\mathcal{D}_u|} \sum_{r \in R}  \sum_{u \in \mathcal{D}_u}\log p(r|u^r)
\end{equation}
where $R$ is the set of four rotation degrees $\{0^{\circ}, 90^{\circ}, 180^{\circ}, 270^{\circ}\}$,
$p(r|u^r)$ is the conditional probability of rotation degree, and $u^r$ is the unlabeled image $u$ rotated by $r$.
This objective function conducts 4-class classification.

Following the authors of MoCo~\cite{moco}, we calculate infoNCE~\cite{InfoNCE}.
For each mini-batch, we obtain queries $\bm{q}$ and keys $\bm{k}$ as described in \S\ref{sec:self}.
For each pair of a query $q$ and a key $k$, we consider that the key is positive if they derive from the same image, otherwise negative.
MoCo uses a dictionary that contains negative keys.
The dictionary is a queue that enqueues keys from the current mini-batch and dequeues keys from the oldest mini-batch.
The dictionary size $K$ is generally much larger than the batch size.
For each query $q$, we calculate the object function as follows,
\begin{equation}
\mathcal{L}_q = -\log \frac{\exp(q^\top k_{pos} / \tau)}{\sum_{i=0}^{K}\exp(q^\top k_i / \tau)}
\end{equation}
where $\tau$ is the temperature value, $k_{pos}$ is the positive key.
This object function conducts ($K$+1)-way classification with the softmax function on $K$ negative keys and 1 positive key.

\section{Experiment and Analysis - Details}\label{sup:experiment}
We show more details of our implementation and the extended version of our experiments.

\subsection{Implementation Detail}\label{sup:imple}
In this section, we describe common factors in our experiments.
Specific factors of each experiment are described at the beginning of each experiment.

\PAR{Description of our settings}
Table~\ref{sup-tab:our_setting} shows the description of our main experimental settings.

\begin{table}[t]
    \begin{center}
        \begin{adjustbox}{width=0.975\linewidth}
        \begin{tabular}{@{}ll@{}}
            \toprule
            Setting & Description \\
            \midrule
            Baseline-synth & Model trained on 2 synthetic datasets (MJ+ST)\\
            Baseline-real & Model trained on 11 real datasets (Real-L) \\
            \textit{Aug.} & Best augmentation setting in our experiments \\
            PR & Combination of \textit{Aug.}, PL and RotNet \\
            \bottomrule
        \end{tabular}
        \end{adjustbox}
    \vspace{-2mm}
    \caption{Description of our experimental settings.}
    \label{sup-tab:our_setting}
    \end{center}
    \vspace{-6mm}
\end{table}

\PAR{Training strategy}
Input images are resized into $32\times100$.
We use He's initialization method~\cite{he2015delving} and gradient clipping at magnitude 5. 
The maximum word length is 25.

We use 94 characters for prediction, same with \cite{ASTER}: 26 upper case alphabets, 26 lower case alphabets, 10 digits, and 32 ASCII punctuation marks.
In addition, 3 special tokens are added: ``[PAD]'' for padding, ``[UNK]'' for unknown character, and `` '' for space between characters.
For CTC decoder, ``[CTCblank]'' token is also added. 
For attention mechanism, ``[SOS]'' and ``[EOS]'', which denote the start and end of sequence, are added.

As shown in Table~\ref{sup-tab:split}, the number of training set is imbalance.
To handle data imbalance, we sample the same number of data from each dataset to make a mini-batch.
For example, when we use 11 datasets for training, we sample 12 images (= round(128/11)) per dataset to make a mini-batch.
As a result, the batch size slightly changes depending on the number of datasets for training.
However, the difference is marginal in our experiments and thus we use the balanced mini-batch.
We also use the balanced mini-batch for three unlabeled datasets: 43 images (= round(128/3)) per dataset to make a mini-batch.

\PAR{Evaluation metric}
As described in \S\ref{implementation}, the word-level accuracy is calculated only on the alphabet and digits.
Following common practice~\cite{ASTER}, the accuracy is calculated only on alphabet and digits, after removing non-alphanumeric characters and normalizing alphabet to lower case.

\begin{table*}[t] 
\begin{center}
\begin{adjustbox}{width=0.975\linewidth}
        \begin{tabular}{@{}clccccccccccc|c@{}}
        \toprule
            & & & \multicolumn{4}{c}{Type of training data} & \multicolumn{7}{c}{Dataset name and \# of data}\\ \cmidrule(l){4-7} \cmidrule(l){8-14}
            &  & 
            & \multicolumn{2}{c}{Synthetic data} & \multicolumn{2}{c}{Real data} 
            & IIIT & SVT & IC13 & IC15 & SP & \multicolumn{1}{c}{CT} & Total \\
            
            & Method & Year & MJ+ST & SA & labeled & unlabeled & 3000 & 647 & 1015 & 2077 & 645 & \multicolumn{1}{c}{288} & 7672 \\
        \midrule
\parbox[t]{2mm}{\multirow{15}{*}{\rotatebox[origin=c]{90}{\textbf{Reported results}}}}
& ASTER~\cite{ASTER} & 2018  & $\checkmark$  & & & & 93.4 & 89.5 & 91.8 & 76.1 & 78.5 & 79.5 & 86.4\\
& SAR~\cite{SAR} & 2019 & $\checkmark$ & $\checkmark$ & & & 91.5 & 84.5 & 91.0 & 69.2 & 76.4 & 83.3 & 83.2 \\
& SAR~\cite{SAR} & 2019 & $\checkmark$ & $\checkmark$ & $\checkmark$ & & 95.0 & 91.2 & 94.0 & 78.8 & 86.4 & 89.6 & 89.2 \\
& ESIR~\cite{ESIR} & 2019  & $\checkmark$ &  & & & 93.3 & 90.2 & 91.3 & 76.9 & 79.6 & 83.3 & 86.8 \\
& MaskTextSpotter~\cite{Mask-textspotter}* & 2019 & $\checkmark$ &  & & & 95.3 & 91.8 & \textbf{95.3} & 78.2 & 83.6 & 88.5 & 89.1 \\
& ScRN~\cite{ScRN}* & 2019 & $\checkmark$ &  & & & 94.4 & 88.9 & 93.9 & 78.7 & 80.8 & 87.5 & 88.2 \\
& DAN~\cite{DAN} & 2020  & $\checkmark$ &  & & & 94.3 & 89.2 & 93.9 & 74.5 & 80.0 & 84.4 & 86.9 \\
& TextScanner~\cite{textscanner}* & 2020 & $\checkmark$ &  & & & 93.9 & 90.1 & 92.9 & 79.4 & 84.3 & 83.3 & 88.3 \\
& TextScanner~\cite{textscanner}* & 2020 & $\checkmark$ &  & $\checkmark$ & & \textbf{95.7} & \textbf{92.7} & 94.9 & \textbf{83.5} & 84.8 & 91.6 & \textbf{91.0} \\
& SE-ASTER~\cite{SE-ASTER} & 2020  & $\checkmark$ &   & & & 93.8 & 89.6 & 92.8 & 80.0 & 81.4 & 83.6 & 88.2 \\
& SCATTER~\cite{scatter} & 2020 & $\checkmark$ & $\checkmark$ & & & 93.7 & \textbf{92.7} & 93.9 & 82.2 & \textbf{86.9} & 87.5 & 89.7 \\
& PGT~\cite{PGT-RealSynthMix3} & 2020 & $\checkmark$  & & & $\checkmark$ & 93.5 & 90.7 & 94.0 & 74.6 & 80.1 & 77.8 & 86.5 \\
& RobustScanner~\cite{robustscanner} & 2020  & $\checkmark$  &  & & & 95.3 & 88.1 & 94.8 & 77.1 & 79.5 & 90.3 & 88.2 \\
& RobustScanner~\cite{robustscanner} & 2020  & $\checkmark$  &  & $\checkmark$ & & 95.4 & 89.3 & 94.1 & 79.2 & 82.9 & \textbf{92.4} & 89.2 \\
& PlugNet~\cite{plugnet} & 2020  & $\checkmark$  &  & & & 94.4 & 92.3 & 95.0 & 82.2 & 84.3 & 85.0 & 89.8 \\
\midrule 
\midrule
\parbox[t]{2mm}{\multirow{10}{*}{\rotatebox[origin=c]{90}{\textbf{Our experiment}}}}
& TRBA \\
& ~~~Original~\cite{TRBA} & 2019  & $\checkmark$  & & & & 87.9 & 87.5 & 92.3 & 71.8 & 79.2 & 74.0 & 82.8 \\
& ~~~Baseline-synth  &   & $\checkmark$  & & & & 92.1 & 88.9 & 93.1 & 74.7 & 79.5 & 78.2 & 85.7 \\
& ~~~+ SA &  & $\checkmark$ & $\checkmark$ &  &  & 93.6 & 88.8 & 92.9 & 76.4 & 81.1 & 84.7 & 87.0\\
\arrayrulecolor{lightgray}\cline{2-14}\arrayrulecolor{black}
& ~~~Baseline-real &  & & & $\checkmark$ &  & 93.5 & 87.5 & 92.6 & 76.0 & 78.7 & 86.1 & 86.6 \\
& ~~~+ MJ + ST &  & $\checkmark$ & & $\checkmark$ & & 95.1 & 91.0 & 94.9 & 79.0 & 82.4 & 89.1 & 89.1\\
& ~~~+ MJ + ST + SA &  & $\checkmark$ & $\checkmark$ & $\checkmark$ & & 95.4 & 91.3 & 95.2 & 80.2 & 83.8 & 92.1 & 89.8\\
\arrayrulecolor{lightgray}\cline{2-14}\arrayrulecolor{black}
& ~~~PR &  & & & $\checkmark$ & $\checkmark$ & 94.8 & 91.3 & 94.0 & 80.6 & 82.7 & 88.1 & 89.3\\
& ~~~+ MJ + ST &  & $\checkmark$ & & $\checkmark$ & $\checkmark$ & 95.2 & 92.0 & 94.7 & 81.2 & 84.6 & 88.7 & 90.0\\
& ~~~+ MJ + ST + SA &  & $\checkmark$ & $\checkmark$ & $\checkmark$ & $\checkmark$ & 95.4 & 92.4 & 95.0 & 81.9 & 84.8 & 89.5 & 90.3\\
\bottomrule
    \end{tabular}
    \end{adjustbox}
    \caption{Extended version of Table~\ref{tab:benchmark} in \S\ref{exp:benchmark}.
    We show the results reported in original papers.
    MJ+ST, SA, and ``*'' denote union of MJSynth and SynthText, SynthAdd, and the model that uses character-level labeled data, respectively.
    Real labeled data has various variants as described in \S\ref{sup:benchmark}.
    In each column, top accuracy is shown in \textbf{bold}.
    }
    \label{sup-tab:benchmark-extended}
    \end{center}
\end{table*}

\PAR{Differences from the base code~\cite{TRBA}}
We use the code of the STR benchmark repository as our base code.
The code used in the STR benchmark is different from our settings:
We use
1) All text in ST (7M) while the base code uses only alphanumeric texts in ST (5.5M),
2) Adam~\cite{adam} instead of Adadelta~\cite{zeiler2012adadelta},
3) An one-cycle learning rate scheduling~\cite{super-convergence},
4) Batch size reduced from 192 to 128,
5) Upper/lower case alphabets, digits, and some symbols (94 characters in total) for training while the base code uses lower case alphabets and digits (36 characters in total) for training, 
6) The balanced mini-batch. This is not used for STR benchmark paper.
7) The CTCLoss from the PyTorch library instead of the CTCLoss from Baidu\footnote{https://github.com/baidu-research/warp-ctc}. 
According to the STR benchmark repository\footnote{https://github.com/clovaai/deep-text-recognition-benchmark/pull/209}, the CTCLoss from Baidu has higher accuracy by about 1\% for CRNN than the CTCLoss from PyTorch library.
In addition, 
8) we construct and use our own validation set.
9) We calculate total accuracy on union of SVT, IIIT, IC13, IC15, SP, and CT, except for IC03.
IC03 usually has higher accuracy than other datasets (over 90\% accuracy), and thus excluding IC03 results in lower total accuracy than that of including IC03.

\PAR{Environment}: All experiments are conducted using Pytorch~\cite{PyTorch} on a Tesla V100 GPU with 16GB memory.

\subsection{Comparison to State-of-the-Art Methods}\label{sup:benchmark}
We present the extended version of Table~\ref{tab:benchmark} in \S\ref{exp:benchmark}. 
Table~\ref{tab:benchmark} in \S\ref{exp:benchmark} lists the methods that use only MJ and ST for training, and evaluate six benchmarks: IIIT, SVT, IC13-1015, IC15-2077, SP, and CT.
In addition to Table~\ref{tab:benchmark} in \S\ref{exp:benchmark}, Table~\ref{sup-tab:benchmark-extended} also lists the methods that use SynthAdd (SA)~\cite{SAR} or both synthetic and real data, 

SA is a synthetic dataset generated by Li \etal~\cite{SAR}.
To compensate for the lack of special characters, Li \etal generated SA using the same synthetic engine with ST~\cite{SynthText}.
SA is used in some STR methods~\cite{SAR, scatter}.

Some methods use both synthetic and real data for training~\cite{SAR, textscanner, robustscanner}.
Their models trained on both synthetic and real data show better performance than their models trained only on synthetic data.
However, fairly comparing the three methods~\cite{SAR, textscanner, robustscanner} is difficult since they use the different numbers of real data (50K~\cite{SAR}, 16K~\cite{textscanner}, and 7K~\cite{robustscanner}).
Although they use different real data, we list them in Table~\ref{sup-tab:benchmark-extended}.

Some methods use character-level labels~\cite{Mask-textspotter,ScRN,textscanner}.
ST has character-level labels, and the researchers use them.
The methods using character-level information tend to obtain higher accuracy on irregular texts.

TRBA with our best setting (TRBA-PR) trained on both synthetic (MJ+ST) and real data has a competitive performance of 90.0\% to state-of-the-art methods.
Adding synthetic data SA for training further improves by 0.3\% (from 90.0\% to 90.3\%).

\begin{figure*}[t]
    \includegraphics[width=0.975\linewidth]{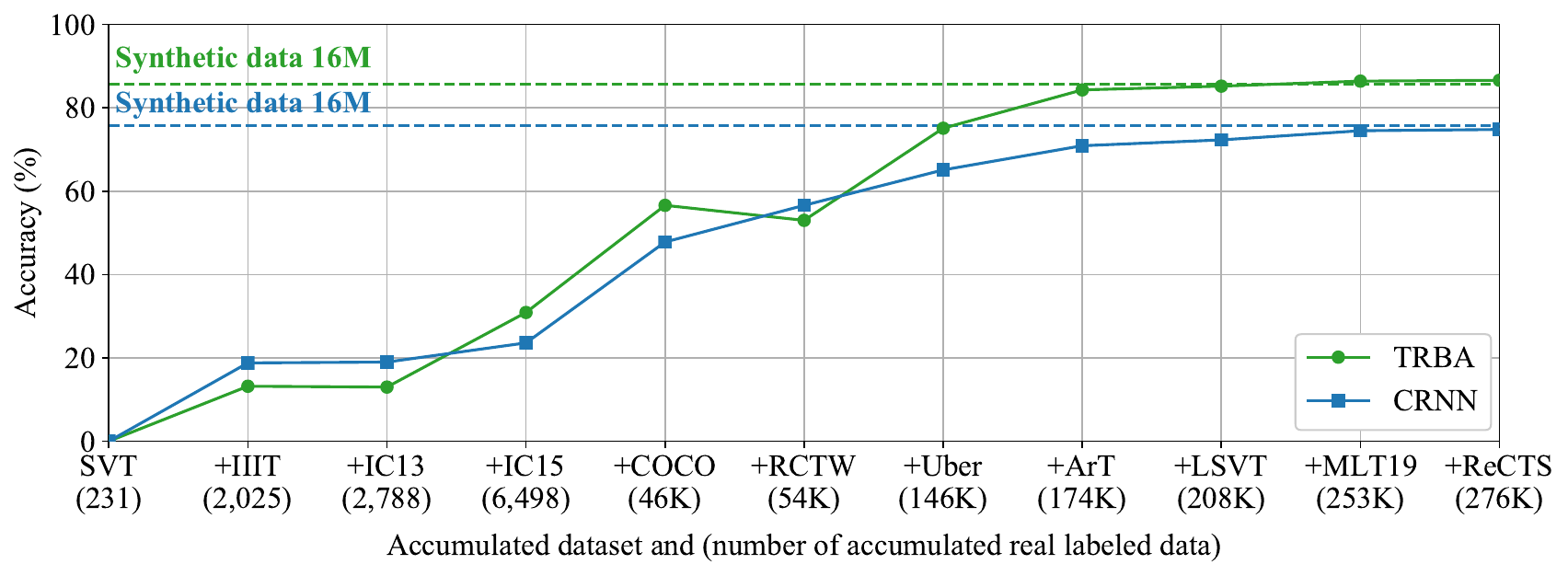}
    \centering
    \vspace{-3mm}
    \caption{Accuracy vs. number of accumulated real labeled data, extended version of Figure~\ref{fig:data_increment} in \S\ref{sec:introduction}.
    Along with the increment of real data, the accuracy obtained using real data approaches the accuracy obtained using synthetic data.
    }
    \label{sup-fig:data_increment}
\end{figure*}

\begin{table}[t] 
    \begin{center}
        \begin{threeparttable}
        \begin{tabular}{@{}lrr@{}}
            \toprule
            Dataset & CRNN & TRBA\\
            \midrule
            SVT\tnote{(a)} & 0.0 & 0.2 \\
            + IIIT & 18.8 & 13.2 \\
            + IC13\tnote{(b)} & 19.0 & 13.0 \\
            + IC15\tnote{(c)} & 23.6 & 30.9 \\ 
            + COCO & 47.8 & 56.6 \\ 
            + RCTW & 56.6 & 53.0 \\
            + Uber\tnote{(d)} & 65.1 & 75.1 \\
            + ArT & 70.9 & 84.3 \\
            + LSVT & 72.3 & 85.2 \\
            + MLT19 & 74.5 & 86.4 \\
            + ReCTS\tnote{(e)} & 74.8 & 86.6 \\
            \bottomrule
        \end{tabular}
        \end{threeparttable}
    \vspace{-2mm}
    \caption{Accuracy vs. dataset increment.
    (a), (b), (c), (d), and (e) denote the results of Year 2011, Year 2013, Year 2015, Year 2017, and Year 2019 in Figure~\ref{fig:data_increment} in \S\ref{sec:introduction}, respectively.
    }\label{sup-tab:data_increment}
    \end{center}
    \vspace{-4mm}
\end{table}

\subsection{Training Only on Real Labeled Data}

\PAR{Accuracy depending on dataset increment: Extended version}
Figure~\ref{sup-fig:data_increment} shows the accuracy vs. the number of accumulated real labeled data per dataset increment.
This is the extended version of Figure~\ref{fig:data_increment} in \S\ref{sec:introduction}.
Table~\ref{sup-tab:data_increment} shows the accuracy along with the increment of real datasets.

\PAR{Improvement by simple data augmentations: Varying degree of each augmentation}
As we mentioned in \S\ref{exp:labeled}, the intensity of each augmentation affects the performance.
We find the best intensities for each augmentation.
Table~\ref{sup-tab:aug_degree} shows the results of them.
We adjust the maximum radius $r$ of Gaussian blur (Blur), the minimum percentage for cropping (Crop), and the maximum degree $^{\circ}$ of rotation (Rot). 

We apply the augmentations with an intensity of 1 to confirm whether applying the weak augmentation will be effective.
In our experiment, the weak augmentations are effective.
1$r$ of Blur improves the accuracy by 0.8\% for CRNN and 0.1\% for TRBA.
99\% of Crop improves the accuracy by 3.0\% for CRNN and 0.5\% for TRBA.
TRBA performs the best at Crop~99\%.
1$^{\circ}$ of Rot improves the accuracy by 0.3\% for CRNN but decreases the accuracy by 0.8\% for TRBA.

CRNN shows the best performance 5$r$ of Blur, 90\% of Crop, and 15$^{\circ}$ of Rot, respectively.
They improve the accuracy by 0.9\%, 4.0\%, and 4.7\% than that of no augmentation, respectively.
TRBA shows the best performance 5$r$ of Blur and 99\% of Crop, respectively. 
They improve the accuracy by 0.2\% and 0.5\% than that of no augmentation, respectively.

When Rot is applied to TRBA, the accuracy of TRBA decreases.
We presume that Rot can hinder the training of TPS in TRBA, which normalizes rotated texts into horizontal texts.
It can result in the decrease in accuracy.
We conduct additional experiments with the model called RBA, which is TRBA without TPS.
When Rot is applied to RBA, the accuracy of RBA increases.
30$^{\circ}$ of Rot improves accuracy by 3.1\% for RBA.
This shows that Rot can improve the accuracy of STR models but might not be compatible with TPS.

In \S\ref{exp:labeled}, we use 15$^{\circ}$ of Rot for TRBA to investigate the effect of combination of augmentations.

\begin{table*}[t]
    \begin{subtable}{0.33\linewidth} \centering
        \begin{tabular}{@{}rrr@{}}
            \toprule
            Radius $r$ & CRNN & TRBA \\
            \midrule
                        0 &  74.8 & 86.6 \\
            1 &  75.6 & 86.7 \\
            5 &  \textbf{75.7} & \textbf{86.8} \\
            10 & 75.4          & 86.7 \\
            15 & 75.0          & 86.5 \\
            \bottomrule
        \end{tabular}
        \caption{Blur}\label{sup-tab:aug_blur}
    \end{subtable}
    \begin{subtable}{0.33\linewidth} \centering
        \begin{tabular}{@{}rrr@{}}
            \toprule
            Percentage \% & CRNN & TRBA \\
            \midrule
                        100   &  74.8 & 86.6 \\
            99    &  77.8 & \textbf{87.1} \\
            95    &  78.6 & 87.0 \\
            90    &  \textbf{78.8} & 86.0 \\
            85    &  78.3          & 85.7 \\
            \bottomrule
        \end{tabular}
        \caption{Crop}\label{sup-tab:aug_crop}
    \end{subtable}
    \begin{subtable}{0.33\linewidth} \centering
        \begin{tabular}{@{}rrrr@{}}
            \toprule
            Degree $^{\circ}$ & CRNN & TRBA & RBA\\
            \midrule
            0     &  74.8 & \textbf{86.6} & 83.8\\
            1     &  75.1 & 85.8 & 83.9\\
            15    &  \textbf{79.5} & 86.2 & 86.7\\
            30    &  79.4 & 86.0 & \textbf{86.9}\\
            45    &  79.3 & 85.6 & 86.5\\
            \bottomrule
        \end{tabular}
        \caption{Rot}\label{sup-tab:aug_rot}
    \end{subtable}
\vspace{-2mm}
\caption{Accuracy with varying degrees of data augmentations.}
\label{sup-tab:aug_degree}
\end{table*}

\subsection{Semi- and self-supervised framework}

\PAR{Details of semi-supervised learning}
For Mean Teacher (MT), we basically follow the code from authors\footnote{https://github.com/CuriousAI/mean-teacher}.
We do not use dropout~\cite{srivastava2014dropout} because most STR models do not use dropout~\cite{CRNN,ASTER,TRBA}.
In our experiments, we use 1.0 for the coefficient $\alpha$ of MT loss.

\PAR{Details of self-supervised learning}
We follow the code from authors\footnote{https://github.com/gidariss/FeatureLearningRotNet}\textsuperscript{,}\footnote{https://github.com/facebookresearch/moco}. 
As a default, we use same hyperparameters with their code except for the number of iterations; we use 200K iterations.
We use the same settings for the pretext task of RotNet:
1) Batch size 512; 128 $\times$ 4 rotations ($0^{\circ}, 90^{\circ}, 180^{\circ}, 270^{\circ}$).
2) SGD optimizer with same setting.
We use the same settings for the pretext task of MoCo:
1) Batch size 256.
2) SGD optimizer with same setting.
3) Same augmentation policy: resized crop, gray scale, color jitter, and horizontal flip.

\subsection{Varying Amount of Real Labeled Data}\label{sup:varying}
To investigate the effect of the amount of real labeled data, we vary the ratio of each dataset while maintaining the diversity of datasets (keep using 11 datasets).
Table~\ref{sup-tab:varying-ratio} shows the number of word boxes used in the experiments varying amount of real labeled data.
Table~\ref{sup-tab:varying-ratio-value} shows the accuracy with varying amount of real labeled data.

\begin{table}[t] 
    \begin{center}
        \begin{tabular}{@{}llrrrrr@{}}
            \toprule
            & & \multicolumn{5}{c}{\# of word boxes per ratio} \\ \cmidrule(l){3-7}
            \multicolumn{2}{l}{Dataset} & 20\% & 40\% & 60\% & 80\% & 100\%\\
            \midrule
            \multicolumn{5}{l}{\textbf{Real labeled datasets (Real-L)}} \\ 
            ~~ & SVT & 46 & 92 & 138 & 184 & 231 \\
            ~~ & IIIT & 358 & 717 & 1,076 & 1,435 & 1,794\\
            ~~ & IC13 & 152 & 305 & 457 & 610 & 763\\
            ~~ & IC15 & 742 & 1,484 & 2,226 & 2,968 & 3,710 \\ 
            ~~ & COCO & 7,868 & 16K & 24K & 31K & 39K\\ 
            ~~ & RCTW & 1,637 & 3,274 & 4,911 & 6,548 & 8,186\\
            ~~ & Uber & 18K & 37K & 55K & 73K & 92K \\
            ~~ & ArT & 5,771 & 12K & 17K & 23K & 29K \\
            ~~ & LSVT & 6,702 & 13K & 20K & 27K & 34K \\
            ~~ & MLT19 & 9,102 & 18K & 27K & 36K & 46K \\
            ~~ & ReCTS & 4,558 & 9,116 & 14K & 18K & 23K \\
            \arrayrulecolor{lightgray}\hline\arrayrulecolor{black}
            ~~ & Total & 55K & 110K & 166K & 221K & 276K\\
            \bottomrule
        \end{tabular}
    \vspace{-2mm}
    \caption{Number of word boxes used in the experiments with varying amount of real labeled data.}
    \label{sup-tab:varying-ratio}
    \end{center}
    \vspace{-6mm}
\end{table}

\begin{table}[t] 
    \begin{center}
        \begin{tabular}{@{}llrrrrr@{}}
            \toprule
            & & \multicolumn{5}{c}{Accuracy per ratio} \\ \cmidrule(l){3-7}
            & & 20\% & 40\% & 60\% & 80\% & 100\% \\
            \multicolumn{2}{l}{Dataset} & 55K & 110K & 166K & 221K & 276K \\
            \midrule
            \multicolumn{4}{l}{TRBA} \\ 
            ~~ & Baseline-real & 51.2 & 78.2 & 83.6 & 85.4 & 86.6 \\
            ~~ & + \textit{Aug.} & 71.2 & 81.9 & 84.7 & 86.3 & 87.5\\
            ~~ & + PR & 80.2 & 85.1 & 87.0 & 88.5 & 89.3 \\
            \midrule
            \multicolumn{4}{l}{CRNN} \\ 
            ~~ & Baseline-real & 53.4 & 66.6 & 70.5 & 73.0 & 74.8 \\
            ~~ & + \textit{Aug.} & 66.4 & 73.6 & 76.3 & 78.6 & 80.0\\
            ~~ & + PR & 76.8 & 80.4 & 81.9 & 82.6 & 83.4\\
            \bottomrule
        \end{tabular}
    \vspace{-2mm}
    \caption{Accuracy vs. amount of real labeled data. 
    These values are plotted in Figure~\ref{fig:varying} in \S\ref{sec:varying}.}
    \label{sup-tab:varying-ratio-value}
    \end{center}
    \vspace{-6mm}
\end{table}

\end{document}